%% file: PaperForReview.tex
\documentclass[10pt,twocolumn,letterpaper]{article}

\usepackage{cvpr}              

\usepackage{graphicx}
\usepackage{amsmath}
\usepackage{amssymb}
\usepackage{algorithm}
\usepackage{algorithmic}
\usepackage{booktabs}
\usepackage{multirow}
\usepackage{array}
\usepackage{booktabs}
\usepackage{makecell}
\usepackage{float}
\usepackage{diagbox}
\usepackage{changepage}
\usepackage{cuted} 

\usepackage[pagebackref,breaklinks,colorlinks]{hyperref}

\usepackage[capitalize]{cleveref}
\crefname{section}{Sec.}{Secs.}
\Crefname{section}{Section}{Sections}
\Crefname{table}{Table}{Tables}

\def\cvprPaperID{464} 
\def\confName{CVPR}
\def\confYear{2023}

\begin{document}

\title{MDQE: Mining Discriminative Query Embeddings to Segment \\ Occluded Instances on Challenging Videos}

 \author{{Minghan Li \qquad Shuai Li \qquad Wangmeng Xiang  \qquad Lei Zhang\thanks{Corresponding author.} } \\
    	Department of Computing, Hong Kong Polytechnic University  \\
    	{\tt\small liminghan0330@gmail.com, \{csshuaili, cswxiang, cslzhang\}@comp.polyu.edu.hk}
    }

\maketitle

\input{sec/0_abstract.tex}
\input{sec/1_introduction.tex}

\input{sec/2_related.tex}

\input{sec/3_Method.tex}

\input{sec/4_results.tex}

\input{sec/5_conclusion.tex}

\newpage
{\small
\bibliographystyle{ieee_fullname}
\bibliography{PaperForReview}
}

\clearpage
\onecolumn
\def\thesection{\Alph{section}}

\vspace{+1cm}
\begin{center}
 {\huge Supplementary Materials}
\end{center}
\vspace{+1cm}

\appendix

In this supplementary file, we provide the following materials:
  \begin{itemize}
    \item Visualization of initial query locations selected by our query initialization ($cf.$ Section 4.2 in the main paper);
    \item Quantitative results with ResNet50 and ResNet101 on YouTube-VIS 2019 ($cf.$ Section 4.3 in the main paper);
    \item Visualization of segmented instance segmentation on challenging videos ($cf.$ Section 4.3 in the main paper).
  \end{itemize}


\section{Visualization of initial query locations}

\cref{fig:query_init_visual} visualizes the initial query locations selected by our query initialization method on challenging videos of OVIS valid set. We see that frame-level queries are well associated in both spatial and temporal directions, even in crowded scenes. 

\section{Quantitative results with ResNet50 and ResNet101 on YouTube-VIS 2019 valid set}
The quantitative results on YouTube-VIS 2019 valid set are reported in \cref{tab:sota_yt19}. We see that the newly developed VIS methods utilizing transformer-based prediction heads can significantly improve the performance to 49.8\% mask AP with ResNet50 backbone and 51.9\% mask AP with ResNet101 backbone.
Our proposed MDQE employs 5-frame clips as input and a deformable-attention decoder, reaching 47.3\% and 47.9\% mask AP with ResNet50 and ResNet101 backbones, respectively. 
The YouTube-VIS 2019 valid set contains mostly short videos. Therefore, methods like SeqFormer \cite{seqformer} and VITA \cite{heo2022vita}, which take the video-in video-out offline inference, can incorporate the temporal information to distinguish the objects that have obvious differences in feature space. This is the reason why they achieve leading mask AP scores on YouTube-VIS 2019 valid set. However, for objects that have similar-looking appearance or heavy occlusions, methods like SeqFormer \cite{seqformer} and VITA \cite{heo2022vita} cannot extract discriminative features to distinguish them accurately, thereby resulting in unsatisfactory results on OVIS data set ( $cf.$ Section 4.3 in the main paper).

\section{Visualization of segmented instance masks}
Figs. \ref{fig:yt21_visual} - \ref{fig:ovis_visual} show the instance masks obtained by the recently proposed top-performing methods with ResNet50 backbone, including IDOL\cite{IDOL} (ECCV 2022), MinVIS\cite{huang2022minvis} (NIPS 2022), VITA\cite{heo2022vita} (NIPS 2022) and our MDQE.

\textbf{YouTube-VIS 2021 valid set.} \cref{fig:yt21_visual} shows the visual comparisons on YouTube-VIS 2021 valid set. We can see that all recent top-performing methods can well process the simple videos on YouTube-VIS 2021 valid set.

\textbf{OVIS valid set.} \cref{fig:ovis_visual} and \cref{fig:ovis_visual2} compare the visual results of instance segmentation on OVIS valid set. Compared with IDOL \cite{IDOL} and MinVIS \cite{huang2022minvis} with per-frame input, our MDQE with per-clip input can exploit richer spatio-temporal features of objects, thereby segmenting occluded instances better. On the other hand, MDQE with discriminate query embeddings can track instances with complex trajectories more accurately in challenging videos, such as cross-over objects and heavily occluded objects in crowded scenes.

\input{fig/query_init_visual.tex}
\input{tab/sota_yt19.tex}

\input{fig/yt21_visual.tex}
\input{fig/ovis_visual.tex}
\input{fig/ovis_visual2.tex}  

\end{document}

%% file: sec/0_abstract.tex
\begin{abstract}

While impressive progress has been achieved, video instance segmentation (VIS) methods with per-clip input often fail on challenging videos with occluded objects and crowded scenes. This is mainly because instance queries in these methods cannot encode well the discriminative embeddings of instances, making the query-based segmenter difficult to distinguish those `hard' instances. To address these issues, we propose to mine discriminative query embeddings (MDQE) to segment occluded instances on challenging videos.
First, we initialize the positional embeddings and content features of object queries by considering their spatial contextual information and the inter-frame object motion.
Second, we propose an inter-instance mask repulsion loss to distance each instance from its nearby non-target instances. 
The proposed MDQE is the first VIS method with per-clip input that achieves state-of-the-art results on challenging videos and competitive performance on simple videos. In specific, MDQE with ResNet50 achieves 33.0\% and 44.5\% mask AP on OVIS and YouTube-VIS 2021, respectively. 
Code of MDQE can be found at \url{https://github.com/MinghanLi/MDQE_CVPR2023}.

\end{abstract}

%% file: sec/1_introduction.tex
\section{Introduction}
\label{sec:intro}

\input{fig/overview.tex}

Video instance segmentation (VIS) \cite{yang2019video} aims to obtain pixel-level segmentation masks for instances of different classes over the entire video.
The current VIS methods can be roughly divided into two paradigms: per-frame input based methods \cite{yang2019video, bertasius2020classifying, Li_2021_CVPR, yang2021crossover, ke2021pcan, IDOL, huang2022minvis} and per-clip input based methods \cite{Athar_Mahadevan20stemseg, li2022cico, yang2022TempEffi, wang2020vistr, wu2022trackletquery, hwang2021video, seqformer}. 
The former paradigm first partitions the whole video into individual frames to segment objects frame by frame, and then associate the predicted instance masks across frames, while the latter takes per-clip spatio-temporal features as input to predict multi-frame instance masks with the help of embedding learning \cite{Athar_Mahadevan20stemseg}, graph neural networks \cite{wang2021end} and transformer networks \cite{wang2020vistr, hwang2021video, seqformer,heo2022vita}. 

The recently proposed per-clip VIS methods \cite{wang2020vistr,wu2022trackletquery,hwang2021video,seqformer} have set new records on the YouTube-VIS datasets \cite{yang2019video}, achieving significant performance improvement over the per-frame VIS methods \cite{cao2020sipmask,Li_2021_CVPR, yang2021crossover,QueryInst,bertasius2020classifying, zhu2022IAI, he2023inspro}.  
SeqFormer \cite{seqformer} and VITA \cite{heo2022vita} locate an instance in each frame and aggregate temporal information to learn powerful representations of video-level instances via a naive weighted manner and a video-level decoder, respectively.
However, on the challenging OVIS dataset \cite{qi2021occluded}, which includes occluded or similar-looking instances in crowded scenes, the per-clip VIS methods lag behind the per-frame ones. 
Actually, the recently developed per-frame method IDOL\cite{IDOL} records state-of-the-art performance on OVIS by introducing contrastive learning \cite{pang2021quasi, wang2021densectt, dave2022temporalctt} to learn inter-frame instance embeddings. We argue that the per-clip VIS methods should be able to exploit richer spatial-temporal features and achieve better performance than their per-frame counterparts. However, there are two main issues that limit the existing per-clip methods to achieve this goal.

First, existing query-based VIS methods adopt zero or random input as the positional embeddings and content features of object queries in decoder layers, which cannot encode spatio-temporal prior of objects, resulting in poor results on challenging videos. 
Second, during training, the existing mask prediction loss mainly forces each query to match the pixels of its target instance \cite{Wang2022InstUnique,ke2021occlusion-aware} and mismatch the pixels of other instances and the background. No further inter-instance clue has been exploited to teach the segmenter to distinguish mutually occluded instances. 

To address the above issues, we propose to mine discriminative query embeddings (MDQE) to better segment hard instances on challenging videos for per-clip VIS methods. First, we propose to improve the initialization of object queries to specify discriminative spatial-temporal priors. We divide the activation map of each frame into several patches via a grid and select the peak point in each patch as the initial positions of frame-level queries, and then associate them across frames by embedding similarity to ensure that frame-level queries in the same grid of the video clip can correspond to the same object.
Second, to teach the query-based segmenter to distinguish occluded instances, we replace the original mask prediction loss with an inter-instance mask repulsion loss, which forces each query to activate the pixels of its target instance and suppress the pixels of its surrounding non-target instances. 

The proposed VIS method with per-clip input, namely MDQE, is the first to achieve contrastive learning of instance embeddings via query initialization and the inter-instance mask repulsion supervision, which can effectively segment hard instances on challenging videos. 
Our experiments on both OVIS and YouTube-VIS datasets validate that MDQE with per-clip input achieve competitive performance with its per-frame competitors. 

%% file: fig/overview.tex
\begin{figure*}
\begin{center}
\includegraphics[width=0.98\textwidth]{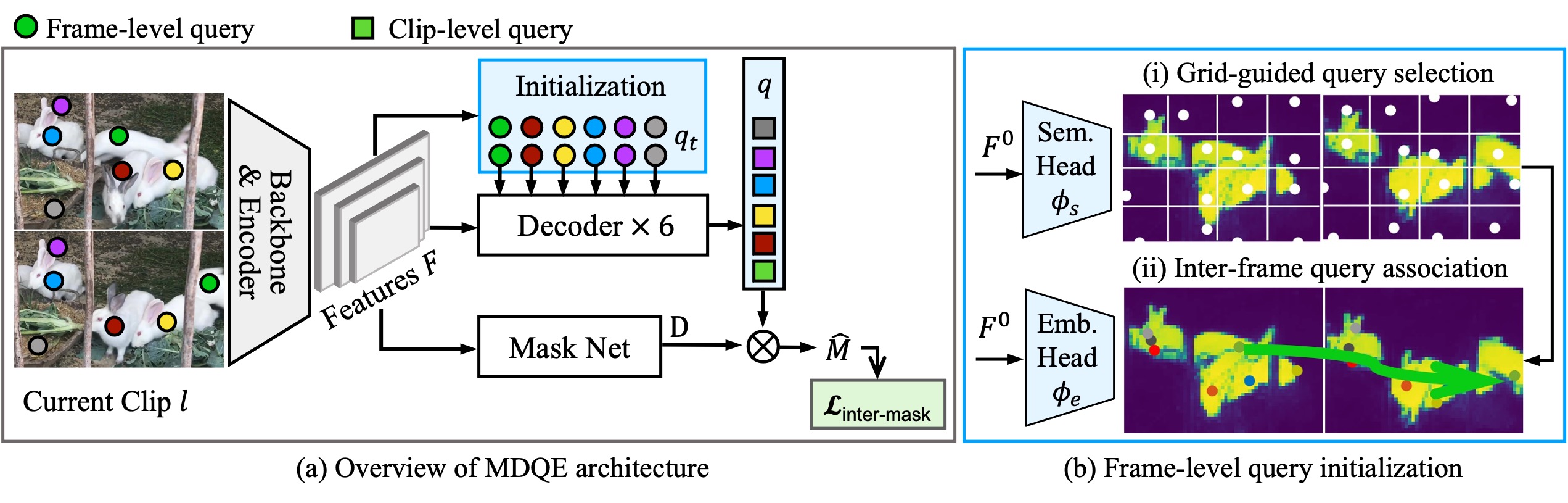}
\end{center}
\vspace{-7.5mm}
\caption{
(a) The proposed MDQE architecture consists of a backbone and encoder that extract multi-scale features $F$ from a video clip, a query initialization module that produces temporally-aligned frame-level queries $q_t$, a decoder that decodes discriminative clip-level queries $q$, and a Mask Net that generates mask features $D$.
The mask features $D$ and clip-level queries $q$ are combined via a linear combination to obtain the clip-level instance mask $\hat{M}$, which is supervised by our proposed inter-instance mask repulsion loss in \cref{sec:inter-inst}.
(b) The frame-level query initialization consists of two steps: grid-guided query selection and inter-frame query association, resulting in temporally-aligned frame-level queries. Please refer to \cref{sec:query-init} for more details. 
}\label{fig:overview}
\vspace{-3mm}
\end{figure*}

%% file: sec/2_related.tex
\section{Related work}
\label{sec:related}
Our work is related to the many per-frame VIS methods, the recently proposed per-clip VIS methods, as well as the methods for learning query embeddings.

\textbf{Per-frame input based VIS methods.} A popular VIS pipeline \cite{yang2019video, cao2020sipmask,Li_2021_CVPR,lin2021video,liu2021sg,yang2021crossover,QueryInst,ke2021pcan,huang2022minvis,IDOL} is to extend the representative image instance segmentation methods \cite{he2017mask,bolya2019yolact,tian2020conditional,carion2020end,cheng2021mask2former} by adapting a frame-to-frame instance tracker. For example, in \cite{yang2019video,QueryInst,huang2022minvis}, the clues such as category score, box/mask IoU and instance embedding similarity are integrated into the tracker. However, these trackers may struggle in distinguishing instances with similar appearance. Inspired by contrastive learning \cite{chen2020simplectt, pang2021quasi,khosla2020supervisedctt,wang2021densectt,dave2022temporalctt}, IDOL \cite{IDOL} learns discriminative instance embeddings for multiple object tracking frame by frame, achieving state-of-the-art results on OVIS \cite{qi2021occluded}.
Besides, clip-to-clip trackers \cite{bertasius2020classifying, lin2021video, qi2021occluded} propagates the predicted instance masks from a key frame to other frames using deformable convolution \cite{bertasius2020classifying,dai2017deformable}, non-local block \cite{lin2021video}, correlation \cite{Li_2021_CVPR,qi2021occluded}, graph neural network \cite{wang2021end}, \etc. By exploiting the temporal redundancy among overlapped frames, clip-to-clip trackers improve much the performance of per-frame methods.

\textbf{Per-clip input based VIS methods.} A clip-in clip-out VIS pipeline was firstly proposed in \cite{Athar_Mahadevan20stemseg} to model a video clip as a single 3D spatio-temporal pixel embedding. 
In recent years, transformer based per-clip methods \cite{wang2020vistr,wu2022trackletquery,hwang2021video,yang2022TempEffi,seqformer} have achieved impressive progress on the YouTube-VIS datasets \cite{yang2019video}. VisTR \cite{wang2020vistr} views the VIS task as a direct end-to-end parallel sequence prediction problem, but it needs a large memory to store spatio-temporal features. To solve the issue, IFC\cite{hwang2021video} transfers inter-frame information via efficient memory tokens, and SeqFormer \cite{seqformer} locates an instance in each frame and aggregates temporal information to predict video-level instances. To keep object temporal consistency, EfficientVIS \cite{wu2022trackletquery} transfers inter-clip query embeddings via temporal dynamic convolution.

However, per-clip VIS methods do not perform well on the challenging OVIS videos \cite{qi2021occluded} with occluded objects in crowded scenes. Actually, occlusion-aware models have been developed for related tasks \cite{wang2018repulsion,xie2021pscocc,ke2021occlusion-aware,miao2021occreid}. For instance, a bilayer convolutional network is developed in \cite{ke2021bcnet} to infer the occluder and occluded instances in image segmentation. A repulsion detection loss is designed in \cite{wang2018repulsion} to distance the bounding box of an object from the surrounding non-target objects for detecting individual pedestrian in a crowd. Inspired by these works, we propose an inter-instance mask repulsion loss to distinguish the pixels of each instance from its nearby non-target instances.

\textbf{Query initialization.} Existing query-based VIS methods adopt zero-initialized (\eg, DETR \cite{carion2020end}) or randomly-initialized (\eg, Deformable DETR\cite{zhu2020deformable}) inputs as initial queries. The initial queries cannot encode well the spatio-temporal priors of objects, making the query-based segmenter difficult to distinguish occluded instances with similar appearance. Actually, query initialization with contextual and positional information has been used in many computer vision tasks \cite{li2022maskdino, bai2022lidatransfusion,he2022queryprop, pei2022osformer} for higher performance or faster convergence. 
However, it has not been well explored in VIS. In this paper, we thus propose a query initialization method to obtain temporally-aligned frame-level queries.

%% file: sec/3_Method.tex
\section{Methodology}\label{sec:method}
We outline the proposed MDQE from the perspective of query-based mask prediction in \cref{sec:overview}, then introduce the two major parts of MDQE: object query initialization in \cref{sec:query-init} and the inter-instance mask repulsion loss in \cref{sec:inter-inst}. Finally, we present the training loss and near-online inference process in \cref{sec:entire_arch}.

\subsection{Framework Overview}\label{sec:overview}
An input video is partitioned into a few video clips, each with $T$ frames.
As illustrated in the left of \cref{fig:overview}, during training, a video clip first passes through the backbone and encoder to obtain its multi-scale features $F$, where the one at the largest scale is represented as $F^0$.
The feature $F^0$ is then used as the input to our proposed query initialization module in \cref{sec:query-init} to produce temporally-aligned frame-level queries $\{q_t\}_{t=1}^T$ (denoted by circles), which will be fed into the decoder to obtain discriminative clip-level queries $q$ (denoted by squares). 
On the other hand, the multi-scale features $F$ are input into the Mask Net to generate mask features $D$, which are then combined with the clip-level queries $q$ to produce clip-level instance masks $\hat{M}$ using linear combination \cite{bolya2019yolact, tian2020conditional}. Finally, the predicted masks $\hat{M}$ are supervised by our proposed inter-instance mask repulsion loss in \cref{sec:inter-inst} to distance each instance from its nearby non-target instances, implementing contrastive learning of instance masks.

\subsection{Frame-level Query Initialization}\label{sec:query-init} 
In this subsection, we initialize frame-level queries with instance-specific information and improve the decoder architecture to mine discriminative instance embeddings.

\input{fig/dec_arch.tex}

Existing query-based VIS methods typically adopt zero or random input as the initial positional embeddings and content features of object queries, following DETR \cite{carion2020end} and deformable DETR \cite{zhu2020deformable}. 
However, a recent method \cite{pei2022osformer} called \textit{grid-guided query selection} was proposed to endow object queries with positional and contextual information of the input image, as shown in the top of \cref{fig:overview}(b). This method inputs the features $F^0$ into a semantic segmentation head ($\phi_s$ with three linear layers) to predict the class-aware activation map: $S=\phi_s(F^0) \in R^{c\times T\times H^0\times W^0}$, where $c$ is the number of categories and $H^0, W^0$ are the height and width of features. The activation map is evenly divided into several patches through a grid, and the peak point with the highest response (the white dots) in each patch is selected. The coordinates $p \in R^2$ and features $F^0_p \in R^d$ of the peak point are then assigned as the initial positions and content features of the query, where $d$ is the dimension of features.

However, the grid-guided query selection may not be able to cover an object with the same grid across multiple frames because object motion or camera jitters will cause position shifts over time. To improve temporal consistency, we extend the grid-guided query selection by incorporating the \textit{inter-frame query association}, as illustrated in the bottom of \cref{fig:overview}(b). 
For a video clip, we first perform the above grid-guided query selection frame by frame to obtain the initial frame-level queries, and then input the content features of these queries into an embedding head ( $\phi_e$ with three linear layers) and output their embeddings: $e_p=\phi_e(F^0_p)\in R^{d_e}$, where $p$ represents the query position and $d_e$ is the dimension of embeddings. 
To obtain temporally-aligned queries, we calculate the embedding similarity between each query in the central frame and neighboring queries within a large window in the non-central frame. The query in the non-central frame with the highest similarity is assigned as the matched query. Note that the size of the window increases if two frames are far apart. After applying the above inter-frame query association, frame-level queries within the same grid are roughly aligned across multiple frames.

In training, we employ the commonly used focal loss \cite{lin2017focal} to supervise the class-aware activation map, denoted as $\mathcal{L}_\textbf{init-sem}$.
Besides, we employ contrastive learning \cite{pang2021quasi,IDOL} to pull the query embeddings with the same instance ID over multiple frames closer, and push them away from each other otherwise. For an object query at position $p$, its contrastive loss of embeddings is defined as:
\begin{align}
    \mathcal{L}_\textbf{init-reid} &= -\log \frac{\exp(e_p \cdot e^{r+}_{p'})}{\exp(e_p \cdot e^{r+}_{p'}) + \sum_{r-}\exp(e_{p} \cdot e^{r-}_{p'})}, \nonumber
\end{align}
where $e_p$ represents its query embeddings, $e^{r+}_{p'} $ and $e^{r-}_{p'}$ denote the embeddings vectors of its neighbouring queries at position $p'$ with the same instance ID and with different instance IDs, respectively. 

Since our query initialization can provide instance-specific spatial-temporal information for frame-level queries, we further adjust the decoder architecture to take full advantage of it. We compare the architectures of the first decoder layer of SeqFormer\cite{seqformer} and our MDQE in \cref{fig:dec_arch}. As shown in \cref{fig:dec_arch}(a), SeqFormer employs random input as clip-level queries, calculates cross-attention between clip-level queries and per-frame encoded features to specify frame-level queries in each frame, and finally updates clip-level queries as the weighted combination of frame-level queries. We refer to this architecture of coping embeddings from clip-level instance queries to frame-level object queries as `I2O'.

Our MDQE utilizes a different decoder architecture, as shown in \cref{fig:dec_arch}(b). it first computes cross-attention and then self-attention, and integrates frame-level object queries into clip-level instance queries (`O2I' for short). 
The clip-level queries $q \in R^{N\times d} $ are calculated similarly to SeqFormer \cite{seqformer}:
$q  = \sum\nolimits_{t=1}^{T} w_t \cdot q_t, $
where $q_t \in R^{N\times d}$ indicates frame-level queries, and $w_t = \text{FFN}(q_t) \in R^{N\times 1}$ is the time weight of each frame in the clip. 
Additionally, we add an extra temporal cross-attention (TCA) layer to mine and integrate discriminative features of instances in a larger spatio-temporal receptive field. The attention module in TCA extracts deformable points from multi-frame single-scale feature maps \cite{zhu2020deformable}.
Our proposed query initialization can early associate instances across frames, providing a good warm-up to the decoder with the `O2I' architecture. This helps to reduce confusing masks among crowded instances.

\subsection{Inter-instance Mask Repulsion Loss }\label{sec:inter-inst}
\input{fig/inter_mask}
The clip-level instance masks $\hat{M} \in R ^{N\times T\times H\times W}$ can be produced by the linear combination of mask features $D\in R^{d \times T\times H\times W}$ and clip-level queries $q \in R^{N\times d}$:
    $\hat{M} = q D,$
as shown in \cref{fig:overview}(a).
During training, the predicted instance masks $\hat{M}$ are supervised by the ground-truth instance masks $M$ via the binary cross-entropy (BCE) loss and the Dice loss \cite{dice1945dice,milletari2016dice_vnet}.
The typical instance mask prediction loss thus can be formulated as: 
\begin{equation}
    \mathcal{L}_{\textbf{mask}} = \mathcal{L}_\textbf{BCE}(\hat{M},\ M) + \mathcal{L}_\textbf{Dice}(\hat{M},\ M),
    \label{eq:intra-inst}
\end{equation}
where $M \in R^{K\times T\times H\times W}$ is the ground-truth (GT) instance mask, and $K$ is the number of matched GT instances. The formula of Dice loss is as follows:
\begin{equation}\label{eq:dice}
    \mathcal{L}_\textbf{Dice}(\hat{M},\ M) = \frac{1}{K} \sum\nolimits_{i=1}^K 1-\frac{2|\hat{M_i} \odot M_i|}{|\hat{M_i}|+|M_i|}
\end{equation}
where $\odot$ denotes the point-wise matrix multiplication and $|\cdot|$ sums the values in the matrix.

However, the mask prediction supervision gives priority to match the pixels of its target instance, and then mismatch the pixels of other instances and the background. This may make the query-based segmenter converge to a shortcut of object spatial locations, resulting in imprecise segmentation prediction and confusing masks on occluded instances.

In fact, the relative relationship between each instance and its surrounding instances can provide contrastive features to the query-based segmenter. Inspired by the success of contrastive learning \cite{wang2021densectt,chen2020simplectt,ke2021occlusion-aware}, we design an inter-instance mask repulsion loss to suppress the pixels belonging to the nearby non-target instances. 
For the $i$-th instance, we define its nearby non-target instances via the intersection of union of bounding boxes (BIoU):
\begin{equation}\label{eq:biou-inter}
    o_i = \{ j\ | \max_{t\in [1, T]} \text{IoU}(B_{ti}, B_{tj})\! >\! \epsilon,\forall j\! \in\! [1, K], j \!\neq\! i\}, 
\end{equation}
where $\epsilon $ is a threshold to control the number of nearby non-target samples, which is set to 0.1 by default. Let the union of nearby GT instance masks be the complementary GT inter-instance mask, \ie, $M_{o_i} = \cup_{j\in o_i} M_j$,
which contains all pixels of its nearby non-target instances, as illustrated in \cref{fig:inter_mask}. Since most annotations used in instance segmentation are not mutually exclusive, we further set $M_{o_i} = M_{o_i} \cap (1-M_i)$ to remove pixels contained in the GT mask and the GT inter-instance mask at the same time.

Typically, the supervision for predicting instance masks includes the BCE and Dice losses in \cref{eq:intra-inst}. In order to train the segmenter to perceive the relative relationships between each instance and its surroundings, we enhance the original BCE and Dice losses by incorporating inter-instance repulsion supervision, named inter-instance BCE loss and inter-instance Dice loss, respectively.
Specifically, we adopt a weighted BCE loss to assign a larger weight for the pixels belonging to the target instance and its nearby instances. The formula of the inter-instance BCE loss is:
\begin{align}
    \mathcal{L}_\textbf{BCE-inter}  = \frac{1}{|W_i|} \sum\nolimits_{p=1}^N  W_{ip}\ \text{BCE}(\hat{M}_{ip}, M_{ip}), 
    \label{eq:inter-inst}
\end{align}
where $p$ and $N$ indicate the pixel position index and the total number of points in the mask, respectively. The corresponding inter-instance weight $W_{ip}$ is set to $\alpha$ (2 by default), if $M_{ip} = 1$ or $M_{{o_i}p} = 1$, otherwise 1.

On the other hand, we introduce an inter-instance mask repulsion loss, which involves $M_{o_i}$ into the Dice loss to explicitly suppress the pixels of nearby non-target instances. The formula of inter-instance Dice loss is:
\begin{align}
    \mathcal{L}_\textbf{Dice-inter}
     = 1 - \frac{2|\hat{M}_i \odot M_i| \ + \ |(1-\hat{M}_i) \odot M_{o_i}|}{|\hat{M}_i|\ +\ |M_i| \ + \ |M_{o_i}|}. \label{eq:dice-inter}
\end{align}
If an instance is isolated to other instances, \ie, $|M_{o_i}|=0$, the inter-instance Dice loss degrades to the original Dice loss in \cref{eq:dice}. In terms of gradient back-propagation, the pixels that belong to the target instance and its nearby instances will have a larger gradient value compared to other pixels in the image. 

Finally, our inter-instance mask repulsion loss is
\begin{align}
    \mathcal{L}_{\textbf{inter-mask}} = \frac{1}{K} \sum\nolimits_{i=1}^K &\quad \mathcal{L}_\textbf{BCE-inter}(\hat{M}_i, M_i, M_{o_i}) \nonumber \\
    & + \mathcal{L}_\textbf{Dice-inter}(\hat{M}_i, M_i, M_{o_i}).
\end{align}
The above loss considers both the matching of pixels belonging to the target instance and the mismatching of pixels belonging to the nearby non-target instances, therefore providing instance discriminative information to the segmenter for producing more accurate instance masks.


\subsection{Training and Inference Details} \label{sec:entire_arch}
We employ Deformable DETR \cite{zhu2020deformable} as the transformer framework, and SeqFormer \cite{seqformer} as the clip-level VIS baseline. The training losses of our proposed MDQE is:
\begin{equation}
\begin{aligned}
    \mathcal{L}_{total}\ = \ & \lambda_1\ \mathcal{L}_{\textbf{cls}} + \lambda_2\ \mathcal{L}_{\textbf{box}} + \lambda_3\ \mathcal{L}_{\textbf{inter-mask}} \\
    & + \lambda_4\ \mathcal{L}_{\textbf{init-sem}} + \lambda_5\ \mathcal{L}_{\textbf{init-reid}}, \nonumber
\end{aligned}
\vspace{-1mm}
\end{equation}
where we adopt the focal loss for classification, and the smooth $L_1$ loss and the generalized IoU loss \cite{giou} for bounding box regression. During training, we empirically set $\lambda_1 = 2$, $\lambda_2 = 2$, $\lambda_3 = 4$, $\lambda_4=2$, and $\lambda_5=0.5$.

\input{fig/inference.tex}

During inference, MDQE processes the testing video clip by clip in a near-online fashion. 
Multiple frames (more than $T$ frames) are loaded into the backbone and encoder to obtain encoded features, which are then extracted clip by clip to the decoder to output clip-level queries. Overlapping frames between clips are only used in the decoder, making MDQE fast. 
In each clip, instances with classification confidence below a threshold are removed, and their masks and embeddings are added to the memory pool to remember objects from previous clips. 
For a new clip, denoted as the $l$-th video clip for clarity, MDQE generates instance masks $\hat{M}$ and clip-level embeddings $q$ with high classification confidence, and extracts memory-based instance masks $\hat{M}^m$ and embeddings $q^{m}$ from the previous $T_\text{mem}$ clips. 
The Hungarian matching algorithm is applied on the score matrix $S$ to associate instances across clips: 
\begin{equation}\label{eq:tracking}
    S = \beta_1\ \text{mIoU}(\hat{M}^m, \hat{M}) + \beta_2\ \text{sim}(q^{m},\ q),
\end{equation}
where `mIoU' and 'sim' measure the mask IoU of instance masks and embedding similarity of instance queries between the memory pool and the current clip, respectively. $\beta_1$ and $\beta_2$ balance the proportions of the two losses, which are set to 1 by default. This process is illustrated in \cref{fig:inference}.

%% file: fig/dec_arch.tex
\begin{figure}[t]
     \centering
     \includegraphics[width=0.48\textwidth]{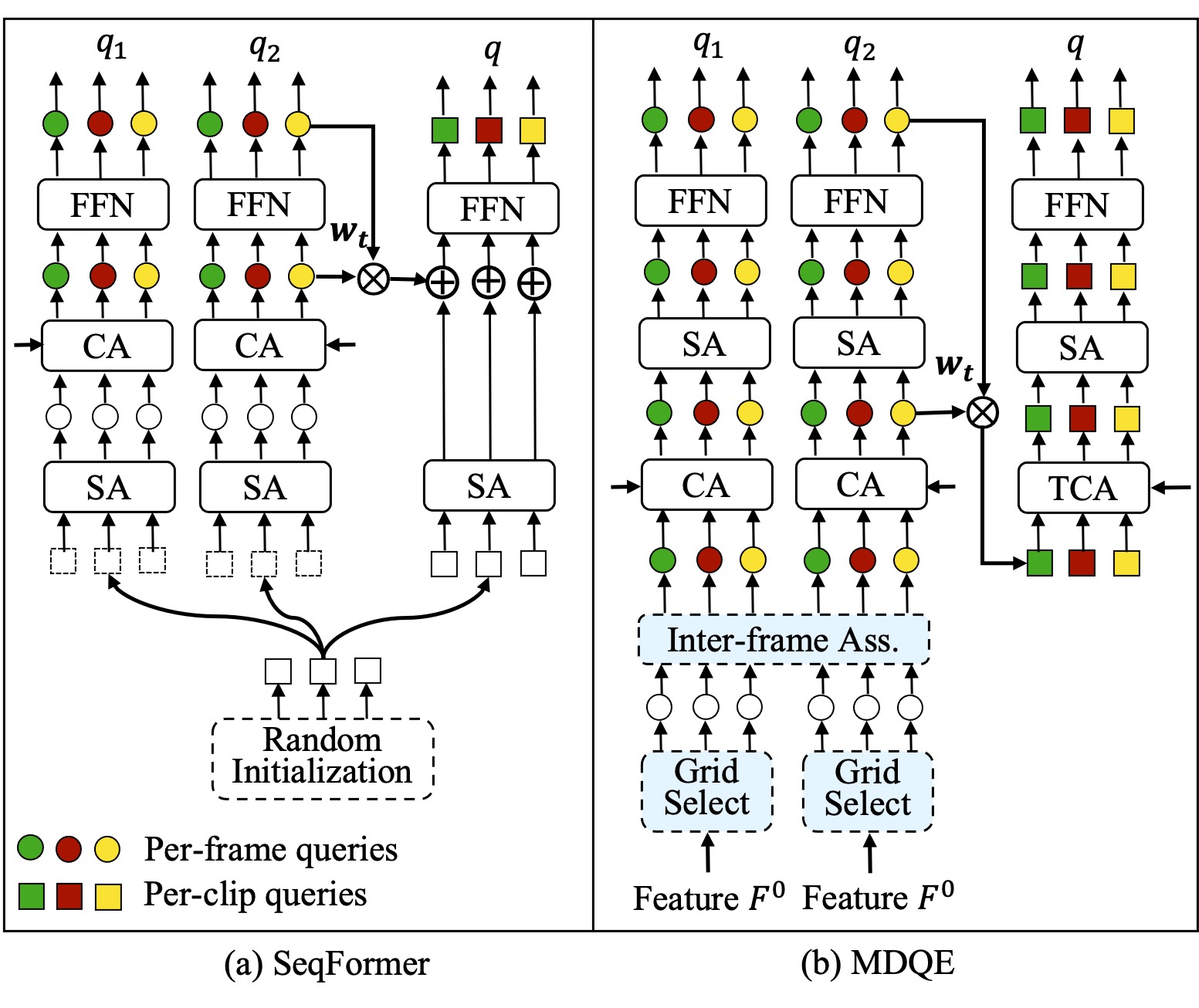}
     \vspace{-8mm}
\caption{Architecture comparison of the first decoder layer between (a) Seqformer\cite{seqformer} and (b) MDQE, where `SA', `CA' and `TCA' refer to self-attention, cross-attention and temporal cross-attention layer, respectively. Please refer to \cref{sec:query-init} for details.}\label{fig:dec_arch}
\vspace{-2mm}
\end{figure}

%% file: fig/inter_mask.tex
\begin{figure}[t]
     \centering
     \includegraphics[width=0.48\textwidth]{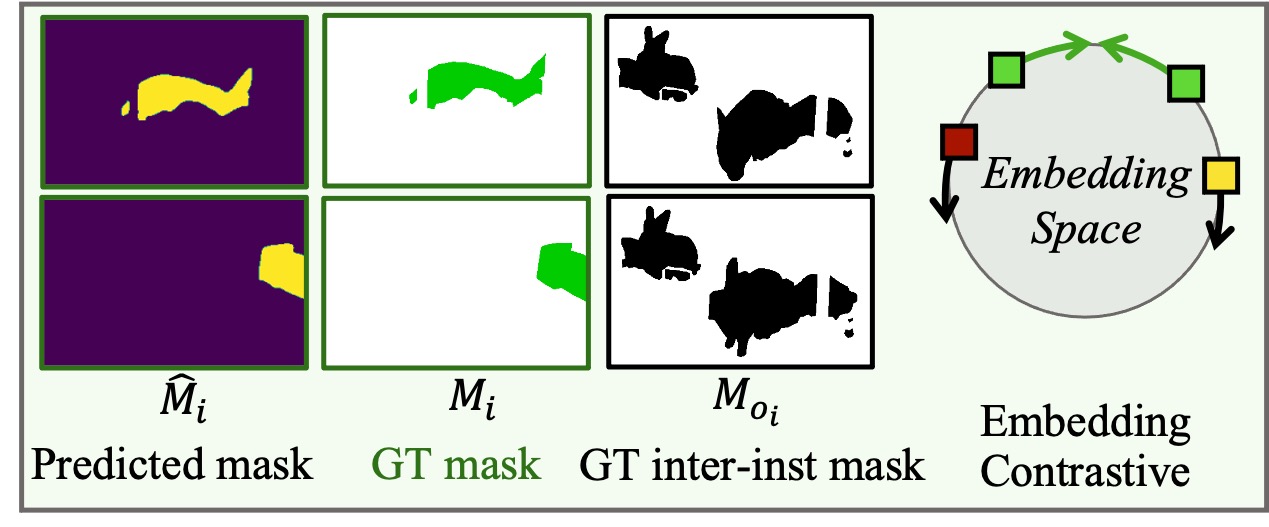}
     \vspace{-6mm}
\caption{Schematic diagram of the inter-inst mask repulsion loss on the instance with ID $i$, corresponding to the instance with green circles in \cref{fig:overview}.}\label{fig:inter_mask}
\vspace{-2mm}
\end{figure}

%% file: fig/inference.tex
\begin{figure}[!t]
     \centering
     \includegraphics[width=0.45\textwidth]{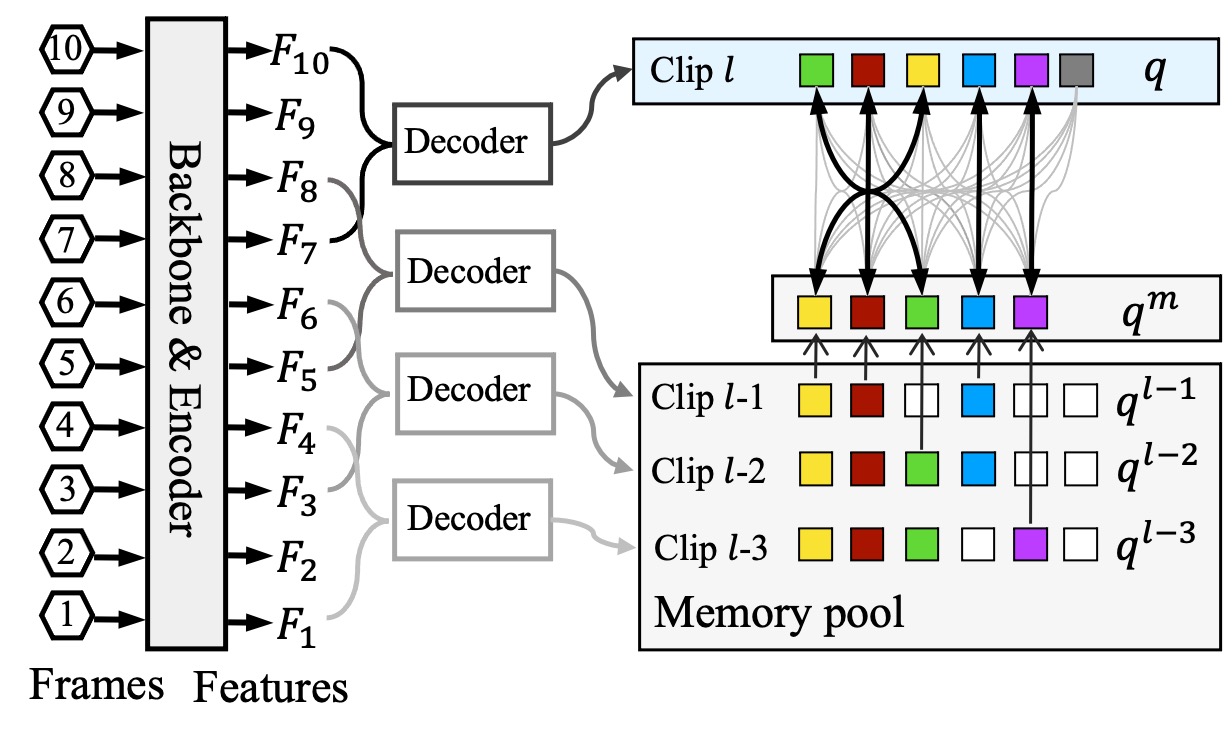}
     \vspace{-3mm}
\caption{Near-online inference with a clip-by-clip tracker.}\label{fig:inference}
\vspace{-2mm}
\end{figure}

%% file: sec/4_results.tex
\input{tab/ablations_ovis.tex}
\input{fig/inter_mask_visual}

\section{Experimental Results}

\subsection{Experiments setup}

\textbf{Datasets.} YouTube-VIS \cite{yang2019video} 2019/2021 respectively contain 2,283/2,985 training, 302/421 validation, and 343/453 test videos over 40 categories. All the videos are annotated for every 5 frames. The number of frames per video is between 19 and 36. 
OVIS \cite{qi2021occluded} includes 607 training, 140 validation and 154 test videos, scoping 25 object categories. Different from YouTube-VIS series, OVIS dataset includes longer videos, up to 292 frames, with more objects per frame and varying occlusion levels. The proportions of objects with no occlusion, slight occlusion, and severe occlusion are 18.2\%, 55.5\%, and 26.3\%, respectively. Note that 80.2\%  of the instances are severely occluded in at least one frame, and only 2\% of the instances are not occluded in any frame. 

\input{tab/sota_yt21_ovis.tex}

\textbf{Evaluation metrics.} The commonly used metrics, including average precision (AP) at different IoU thresholds, average recall (AR) and the mean value of AP (mAP), are adopted for VIS model evaluation. OVIS divides  all  instances  into  three groups called slightly occluded (AP$_\text{so}$), moderately occluded (AP$_\text{mo}$) and heavily occluded (AP$_\text{ho}$), where the occlusion scores of instances are in the range  of [0, 0.25], [0.25, 0.5] and [0.5, 0.75], respectively, whose proportions are 23\%, 44\% and  49\% accordingly. 

\textbf{Implementation details.} Our method is implemented on top of detectron2 \cite{wu2019detectron2}. Following \cite{seqformer}, we set six layers with multi-scale deformable attention module in encoder and decoder, and employ $200$ object queries with dynamic one-to-many label assignment as in \cite{IDOL}.
We first pre-train our model on COCO \cite{lin2014microsoft}, and then finetune it on VIS datasets \cite{yang2019video, qi2021occluded}. In pre-training, we sample an image as the key frame, and resize it in a relative range [0.6, 1] as the reference frame. All models with ResNet50 backbone \cite{he2016deep} are trained on eight GeForce RTX 3090 GPUs with 24G RAM, and models with Swin Large backbone \cite{Liu_2021_ICCV} are trained on four NVIDIA RTX A6000 with 48G RAM. The initial learning rate is 0.0001, and it decays by 0.1 at 20k and 24k iterations. During training and inference, unless specified, all frames are resized to a shorter edge size of $360$ on YouTube-VIS datasets or $480$ on OVIS dataset, and the clip length is taken as $T=4$ on ResNet50 backbone and $T=3$ on Swin Large backbone, respectively.

\subsection{Ablation study}
\vspace{-1mm}
This section performs ablation studies on MDQE and discusses the impact of query embeddings on the segmentation performance. The OVIS valid set is used.

\textbf{Query initialization.} In \cref{tab:abl_init}, we explore the performance improvements brought by query initialization. The baseline Seqformer \cite{seqformer} with `I2O' decoder architecture achieves 15.4\% mask AP. By using grid-guided query selection to initialize frame-level queries, the performance is improved to 19.8\% mask AP. By employing both query initialization and the `O2I' decoder architecture, the mask AP and AP$_{50}$ increase to 24.2\% and 47.5\%, respectively. Additionally, the temporal cross-attention layer (TCA) further improves mask AP by about 1\%. The proposed query initialization significantly enhances instance detection and segmentation accuracy in challenging videos.

\textbf{Inter-frame query association.} In Table \ref{tab:abl_inter_frame_asso}, we investigate the impact of window size $w$ on inter-frame query association for query initialization.
Without inter-frame query association ($w\!=\!0$), the performance is only 28.5\% mask AP. When the window size is set to 3, 5 and 7, the mask AP increases by 1.2\%, 2.1\% and 2.0\%, receptively. So we set the default size as 5. The top row of \cref{fig:inter_mask_visual} visualizes the initial query positions selected by our proposed query initialization method. We see that it keeps temporal consistency and avoids confusion between occluded objects. Visualization of the initial query positions on more videos can be found in the \textbf{supplementary materials}.

\input{fig/abl_study_clip_length}

\textbf{Inter-instance mask repulsion loss.} In \cref{tab:abl_inter_mask}, we compare the performance of typical mask prediction loss and our inter-instance mask repulsion loss.
MDQE with the typical mask prediction loss achieves only 29.0\% mask AP, and performs poorly in all occluded objects.
When the weight $\alpha$ of the inter-instance BCE loss is set to $2$, the performance increases to 30.5\% mask AP, and introducing inter-instance Dice loss further increases mask AP to 31.2\%. 
If we set a higher threshold $\epsilon$ in \cref{eq:biou-inter} to only consider heavily occluded objects, the performance drops by 1.4\% AP$_{so}$ on slightly occluded objects.
Overall, our inter-instance mask repulsion loss brings significant improvements on all occlusion metrics,
3.9\% AP$_{so}$, 3.2\%  AP$_{mo}$ and 1.7\% AP$_{ho}$, respectively, resulting in better instance segmentation results on challenging videos. The visual comparison of instance masks on a video with heavily occluded objects is displayed in \cref{fig:inter_mask_visual}.

\input{tab/sota_occ_metrics.tex}

\textbf{Tracking.} In \cref{tab:abl_track}, we compare the performance of clip-by-clip tracker using different weights in the score matrix in \cref{eq:tracking}. 
Using only the mIou term ($\beta_1\!=\!1$) or the embedding similarity term ($\beta_2\!=\!1$) respectively achieves 29.1\% and 28.3\% mask AP, while the mIou term performs better on moderately and heavily occluded objects. By enabling both the two terms, the mask AP increases to 30.6\% and all occlusion metrics improve significantly. If the number of frames $T_\text{mem}$ in memory pool is reduced from 10 to 5 frames, the performance will drop slightly.

\textbf{Effect of the clip length.} In \cref{fig:abl_clip_length}, we explore the VIS results by varying the clip length. MDQE with $T\!=\!1$ achieves 27\% mask AP, slightly higher than the state-of-the-art per-frame input  method MinVIS\cite{huang2022minvis} in \cref{tab:sota_yt21_ovis}. The mask AP fluctuates between 30.0\% and 31\% with the increase of the frame number of input clip, peaks at around 31.0\% mask AP with 7-frame per clip, and then falls to 29\% mask AP with 9-frame per clip. When $T\!=\!9$, the performance drops because of the complex trajectories of objects in long clips, which can be alleviated by increasing the training samples of long videos.

\subsection{Main Results}
With ResNet50 backbone, we compare  in \cref{tab:sota_yt21_ovis} the proposed MDQE with state-of-the-art methods on OVIS and YouTube-VIS 2021 datasets. For those methods marked by `$^*$', we employ video-in video-out offline inference on YouTube-VIS valid sets (less than 84 frames), and clip-in clip-out inference with overlapping frames between clips on OVIS (at most 292 frames). The occlusion metrics of SOTA methods on OVIS is provided in \cref{tab:sota_occ_matric} as well. 
In addition, we report the performance comparison between recently proposed VIS methods with Swin Large backbone in \cref{tab:sota_swinl}. Due to the limit of space, the experiments on ResNet101 backbone and YouTube-VIS 2019 valid set can be found in the \textbf{supplementary materials}.

\textbf{YouTube-VIS \cite{yang2019video} 2021 valid set.} From \cref{tab:sota_yt21_ovis}, we can see that early VIS methods \cite{yang2019video, cao2020sipmask,Li_2021_CVPR, yang2021crossover} using dense anchors only obtain around 30\% mask AP, while recently proposed VIS methods \cite{IDOL, huang2022minvis,heo2022vita} using sparse object queries can reach more than 44\% mask AP. Since the videos in YouTube-VIS 2021 are short and simple, the performance gap between per-frame input and per-clip input based methods is not significant. Based on the strong decoder layers proposed in Mask2Former, the frame-level method MinVIS \cite{huang2022minvis} and the clip-level method VITA \cite{heo2022vita} respectively reset new state-of-the-arts with 44.2\% and 45.7\% mask AP. Without using the masked-attention in Mask2Former, our proposed MDQE achieves 44.5\% mask AP, which is only slightly lower than VITA \cite{heo2022vita}. 

\input{tab/sota_swinl.tex}

\textbf{OVIS \cite{qi2021occluded} valid set.} OVIS is much more difficult than YouTube-VIS. Its videos have much longer duration with occluded and similar-looking objects. The early per-frame methods MaskTrack R-CNN \cite{yang2019video} and CrossVIS \cite{yang2021crossover} achieve only 10.8\% and 14.9\% mask AP, respectively. The recent per-frame methods with query-based transformer decoder layers, IDOL \cite{IDOL} and MinVIS \cite{huang2022minvis}, bring impressive improvements, achieving 24.3\% and 26.3\% mask AP, respectively.
However, the query-based transformer methods with per-clip input show unexpectedly low performance, such as 15.1\% by Seqformer \cite{seqformer}. By introducing object token association between frames, VITA\cite{heo2022vita} achieves 19.6\% mask AP.
In comparison, our MDQE can reach 29.2\% mask AP, bringing 9.6\% performance improvement. Besides, by using videos of 720p, our MDQE can further improve the mask AP from 29.2\% to 33.0\%, which is the best result using ResNet50 backbone by far. 

We compare the occlusion metrics of competing VIS methods on OVIS valid set in \cref{tab:sota_occ_matric}. One can see that MDQE achieves impressive improvements on AP$_{mo}$ and AP$_{ho}$ metrics, validating that MDQE can handle the moderately and heavily occluded objects very well.


\textbf{Swin Large backbone.} VIS models with the Swin Large backbone can have higher detection and segmentation abilities on challenging videos. Due to limited space, only the recently developed transformer-based methods are compared in \cref{tab:sota_swinl}. IDOL \cite{IDOL}, SeqFormer \cite{seqformer} and our MDQE adopt the deformable DETR transformer architecture, while VITA \cite{heo2022vita} and MinVIS \cite{huang2022minvis} employ the stronger masked-attention transformer architecture.
On both YouTube-VIS 2021 and OVIS valid sets, MDQE obtains competitive performance with VITA \cite{heo2022vita} and MinVIS \cite{huang2022minvis}.
With 720p video input, IDOL \cite{IDOL} with inter-frame object re-association obtains 42.6\% mask AP on OVIS. Due to our limited computational memory, we only take 2-frame video clips as inputs to train our MDQE; however, it can still reach 42.6\% mask AP. We believe MDQE can achieve higher performance if more frames are used in the clip.

\textbf{Parameters and Speed.} We follow Detectron2 \cite{wu2019detectron2} to calculate the parameters and FPS. As shown in \cref{tab:sota_yt21_ovis}, compared with the latest per-clip method VITA \cite{heo2022vita}, our MDQE has 51.4M parameters and runs at 37.8 FPS, saving about 9.4\% parameters and speeding up 12\% the run-time.

The visualization of example segmentation results on challenging videos by the competing VIS methods can be found in the \textbf{supplementary materials}.

%% file: tab/ablations_ovis.tex
\begin{table*}[t]
\begin{subtable}{0.5\textwidth}
    \setlength{\tabcolsep}{0.5mm}{
      \linespread{2}
      \begin{tabular}{p{0.09\textwidth}<{\centering}p{0.1\textwidth}<{\centering}
      p{0.11\textwidth}<{\centering}|p{0.095\textwidth}<{\centering}p{0.095\textwidth}<{\centering}p{0.095\textwidth}<{\centering}p{0.095\textwidth}<{\centering}p{0.095\textwidth}<{\centering}p{0.095\textwidth}<{\centering}}
      \Xhline{0.8pt}
      Init. & Arch. & TCA  & mAP  & AP$_{50}$ & AP$_{75}$  & AP$_{so}$ & AP$_{mo}$ & AP$_{ho}$ \\ 
      \Xhline{0.8pt}
                    & I2O &              & 15.4 & 31.3 & 14.3 & 31.8 & 17.3 & 3.2 \\ 
       $\checkmark$ & I2O &              & 19.8 & 40.6 & 18.2 & 36.3 & 22.6 & 6.5 \\ 
       $\checkmark$ & O2I &              & 24.2 & 47.5 & 22.9 & 40.9 & 27.3 & 8.4 \\ 
       $\checkmark$ & O2I & $\checkmark$ & 25.6 & 49.1 & 24.9 & 41.9 & 29.0 & 11.2 \\ 
     \Xhline{0.8pt}
     \end{tabular}
    }
    \caption{ Initialization for frame-level queries. } \label{tab:abl_init}
\end{subtable}
\begin{subtable}{0.49\textwidth}
    \setlength{\tabcolsep}{0.5mm}{
          \linespread{2}
          \begin{tabular}{p{0.125\textwidth}<{\centering}p{0.165\textwidth}<{\centering}|p{0.095\textwidth}<{\centering}p{0.095\textwidth}<{\centering}p{0.095\textwidth}<{\centering}p{0.095\textwidth}<{\centering}p{0.095\textwidth}<{\centering}p{0.095\textwidth}<{\centering}}
          \Xhline{0.8pt}
          $w$ & Assoc.     & mAP  & AP$_{50}$ & AP$_{75}$  & AP$_{so}$ & AP$_{mo}$ & AP$_{ho}$ \\
          \Xhline{0.8pt}
          0 &                &  28.5 & 53.0 & 26.9 & 47.6 & 32.5 & 11.9 \\
          3 &  $\checkmark$  &  29.7 & 55.6 & 27.1 & 48.9 & 34.5 & 12.2 \\
          5 &  $\checkmark$  &  30.6 & 57.2 & 28.2 & 49.3 & 35.1 & 13.6 \\
          7 &  $\checkmark$  &  30.5 & 57.1 & 28.6 & 49.1 & 33.7 & 13.7 \\
         \Xhline{0.8pt}
         \end{tabular}
    }
    \caption{ Inter-frame query association, where $w$ controls the window size. }\label{tab:abl_inter_frame_asso}
\end{subtable}

\vspace{+0.5mm}

\begin{subtable}{0.5\textwidth}
    \centering
    \setlength{\tabcolsep}{0.5mm}{
      \linespread{2}
      \begin{tabular}{p{0.13\textwidth}<{\centering}p{0.125\textwidth}<{\centering}p{0.07\textwidth}<{\centering}|p{0.092\textwidth}<{\centering}p{0.09\textwidth}<{\centering}p{0.09\textwidth}<{\centering}p{0.092\textwidth}<{\centering}p{0.092\textwidth}<{\centering}p{0.092\textwidth}<{\centering}}
      \Xhline{0.8pt}
       {\small $\mathcal{L}_\textbf{BCE-inter}$} & {\small $\mathcal{L}_\textbf{Dice-inter}$}  & $\epsilon$ & mAP  & AP$_{50}$ & AP$_{75}$  & AP$_{so}$ & AP$_{mo}$ & AP$_{ho}$ \\
      \Xhline{0.8pt}
         &              &     & 29.0 & 51.6 & 29.5 & 44.7 & 31.3 & 11.8 \\
       2 &              & 0.1 & 30.5 & 55.6 & 29.5 & 46.7 & 33.1 & 12.9 \\
       2 & $\checkmark$ & 0.1 & 31.2 & 56.8 & 30.4 & 48.6 & 34.5 & 13.5 \\  
       2 & $\checkmark$ & 0.5 & 30.9 & 56.4 & 30.5 & 47.2 & 34.2 & 13.3 \\
     \Xhline{0.8pt}
     \end{tabular}
    }
    \caption{ Inter-instance mask repulsion loss. }\label{tab:abl_inter_mask}
\end{subtable}
\begin{subtable}{0.49\textwidth}
    \setlength{\tabcolsep}{0.5mm}{
          \linespread{2}
          \begin{tabular}{p{0.09\textwidth}<{\centering}p{0.09\textwidth}<{\centering}p{0.1\textwidth}<{\centering}|p{0.095\textwidth}<{\centering}p{0.095\textwidth}<{\centering}p{0.095\textwidth}<{\centering}p{0.095\textwidth}<{\centering}p{0.095\textwidth}<{\centering}p{0.095\textwidth}<{\centering}}
          \Xhline{0.8pt}
          $\beta_1$ & $\beta_2$ & $T_{\text{mem}}$ & mAP  & AP$_{50}$ & AP$_{75}$  & AP$_{so}$ & AP$_{mo}$ & AP$_{ho}$ \\
          \Xhline{0.8pt}
          1 &   & -  & 29.1 & 54.1 & 27.7 & 46.5 & 32.8 & 12.9 \\  
            & 1 & 10 & 28.3 & 53.4 & 27.1 & 47.1 & 31.3 & 11.6 \\
          1 & 1 & 10 & 30.6 & 57.2 & 28.2 & 49.3 & 35.1 & 13.6 \\
          1 & 1 & 5  & 30.4 & 56.4 & 28.7 & 49.4 & 35.2 & 13.2 \\
         \Xhline{0.8pt}
         \end{tabular}
    }
    \caption{ Tracking. $\beta_1$ and $\beta_2$ control the proportions of mIoU and similarity.}\label{tab:abl_track}
\end{subtable}
\vspace{-3mm}
\caption{Ablation studies of MDQE on OVIS valid set, where the input video is 480p, the clip length is $T=4$, and the number of overlapped frames across clips is $T-1$, respectively. $\epsilon$ controls the number of nearby non-target samples.} \label{tab:abl}
\vspace{-2mm}
\end{table*}

%% file: fig/inter_mask_visual.tex
\begin{figure*}[ht]
     \centering
     \includegraphics[width=0.96\textwidth]{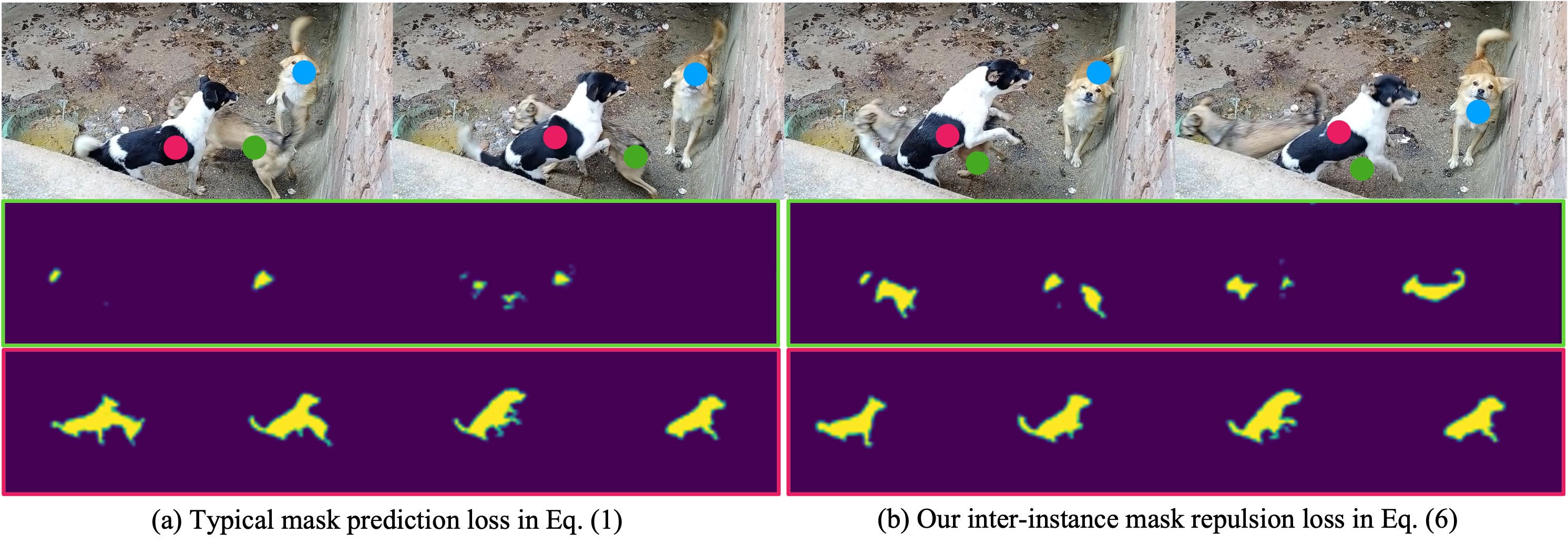}
     \vspace{-4mm}
\caption{Segmentation on heavily occluded objects by MDQE with typical mask prediction loss and our inter-instance mask repulsion loss. The red, green and blue dots highlight the initial query positions selected by our proposed query initialization method. The last two rows display the predicted masks on the occluded instance and the occluder instance.}\label{fig:inter_mask_visual}
\vspace{-3mm}
\end{figure*}

%% file: tab/sota_yt21_ovis.tex
\begin{table*}[t]
\begin{center}
\setlength{\tabcolsep}{0.5mm}{
      \linespread{2}
      \begin{tabular}
      {p{0.11\textwidth}<{\centering}p{0.15\textwidth}|
      p{0.055\textwidth}<{\centering}p{0.05\textwidth}<{\centering}p{0.05\textwidth}<{\centering}
      p{0.05\textwidth}<{\centering}p{0.06\textwidth}<{\centering}|p{0.055\textwidth}<{\centering}p{0.05\textwidth}<{\centering}p{0.05\textwidth}<{\centering}p{0.05\textwidth}<{\centering}p{0.06\textwidth}<{\centering}|p{0.06\textwidth}<{\centering}p{0.06\textwidth}<{\centering}}
         \Xhline{0.8pt}
         \multirow{2}{*}{Type} & \multirow{2}{*}{Methods} & \multicolumn{5}{c|}{YouTube-VIS 2021} & \multicolumn{5}{c|}{OVIS} &  \multirow{2}{*}{FPS}  & \multirow{2}{*}{Params} \\
          & &AP &AP$_{50}$ &AP$_{75}$ &AR$_{1}$ &AR$_{10}$ &AP&AP$_{50}$&AP$_{75}$&AR$_{1}$&AR$_{10}$ &  & \\
         \Xhline{0.6pt}
         \multirow{6}{*}{\shortstack{Per-frame \\ (360p)}}
         &MaskTrack {\small  \cite{yang2019video}}     & 28.6 & 48.9 & 29.6 &- &- & 10.8 & 25.3 & 8.5 & 7.9 & 14.9  & 20.0 & 58.1M\\
         &STMask{\small \cite{Li_2021_CVPR}}                  & 31.1 & 50.4 & 33.5 & 26.9 & 35.6 & 15.4 & 33.9 & 12.5 & 8.9 & 21.4 & 28.0 & -\\      
         &CrossVIS {\small \cite{yang2021crossover}}          & 33.3 & 53.8 & 37.0 & 30.1 & 37.6 & 14.9 & 32.7 & 12.1 & 10.3 & 19.8 & 39.8 & 37.5M\\ 
         & InstFormer{\small \cite{koner2022instanceformer}}  & 40.8  & 62.4 & 43.7 & 36.1 & 48.1 & 20.0 & 40.7 & 18.1 & 12.0 & 27.1 & - & 44.3M\\
         & IDOL {\small \cite{IDOL}}                          & 43.9 & \textbf{68.0} & \textbf{49.6} & 38.0 & 50.9 & 24.3 & 45.1 & 23.3 & 14.1 & \underline{33.2} & 30.6  & 43.1M\\
         & MinVIS {\small \cite{huang2022minvis}}             & 44.2 & 66.0 & 48.1 &\underline{39.2}  &\underline{51.7}  & \underline{26.3} & \underline{47.9} & \underline{25.1} & \textbf{14.6} & {30.0} & 52.4 & 44.0M \\
         
         \Xhline{0.5pt}
         \multirow{6}{*}{\shortstack{Per-clip \\ (360p)}} 
         &VisTR$^*$  {\small \cite{wang2020vistr}}          & 31.8 & 51.7 & 34.5 & 29.7 & 36.9 & 10.2 & 25.7 & 7.7 & 7.0 & 17.4 & 30.0 & 57.2M \\
         &IFC$^*$  {\small \cite{hwang2021video}}           & 36.6 & 57.9 & 39.3 &- &- & 13.1 & 27.8 & 11.6 & 9.4 & 23.9 & 46.5 & 39.3M\\
         & TeViT {\small \cite{yang2022TempEffi}}            & 37.9 & 61.2 & 42.1 & 35.1 & 44.6 & 17.4 & 34.9 & 15.0 & 11.2 & 21.8 & 68.9 & 161.8M\\
         & SeqFromer$^*$ {\small \cite{seqformer}}           & 40.5 & 62.4 & 43.7 & 36.1 & 48.1 & 15.1 & 31.9 & 13.8 & 10.4 & 27.1 & 72.3 &  49.3M\\
         & VITA {\small \cite{heo2022vita}}                   & \textbf{45.7} & \underline{67.4} & \underline{49.5} & \textbf{40.9} & \textbf{53.6} & 19.6 & 41.2 & 17.4 & 11.7 & 26.0 & 33.7 & 57.2M\\
         &MDQE (our)                               & \underline{44.5} & {67.1} & {48.7} & 37.9 & 49.8 & \textbf{29.2} & \textbf{55.2} & \textbf{27.1} & \underline{14.5} & \textbf{34.2} & 37.8 & 51.4M\\
         \Xhline{0.6pt}
         \multirow{2}{*}{720p} 
         & IDOL \cite{IDOL}                         & - & - & - & - & - & 30.2 & 51.3 & 30.0 & 15.0 & 37.5 & - & 43.1M\\
         & MDQE (ours)                              & - & - & - & - & - & \textbf{33.0} & \textbf{57.4} & \textbf{32.2} & \textbf{15.4} & \textbf{38.4} & 13.5 & 51.4M \\
        \Xhline{0.8pt}

      \end{tabular}
}
\end{center}
\vspace{-5mm}
\caption{ Quantitative performance comparison of VIS methods with ResNet50 backbone on benchmark YouTube-VIS 2021 and OVIS datasets. 
Note that MinVIS and VITA adopt stronger masked-attention decoder layers proposed in Mask2Former \cite{cheng2021mask2former}. FPS is computed on YouTube-VIS 2021 valid set, and symbol "-" means the results are not available or applicable. 
Best in \textbf{bold}, second with \underline{underline}.
}\label{tab:sota_yt21_ovis}
\vspace{-3mm}
\end{table*}

%% file: fig/abl_study_clip_length.tex
\begin{figure}[t]
     \centering
     \includegraphics[width=0.42\textwidth]{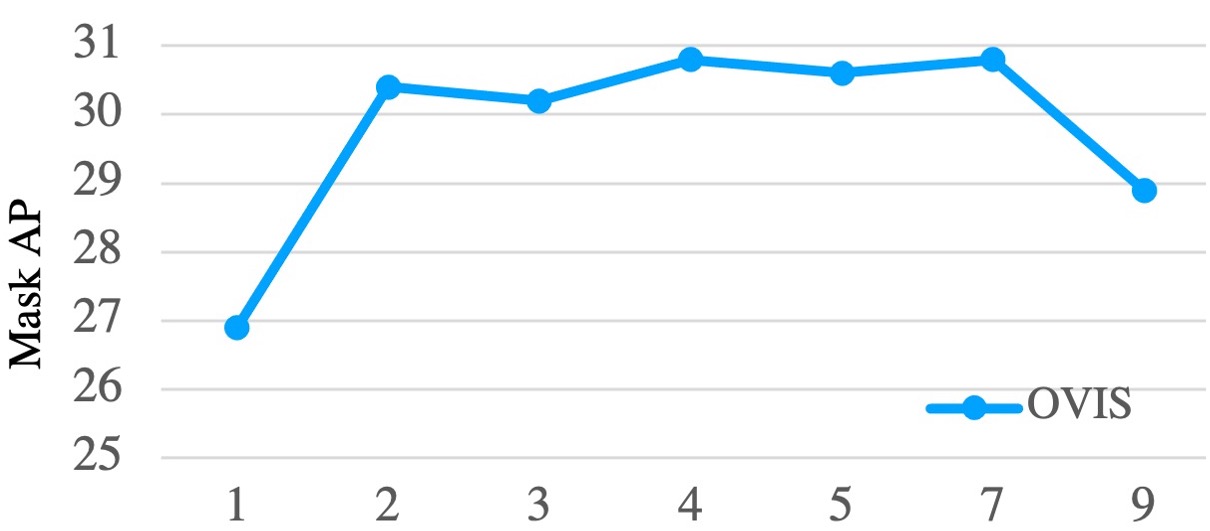}
     \vspace{-2mm}
     \caption{Ablation study on clip length with near-online inference.}
     \label{fig:abl_clip_length}
     \vspace{-3mm}
\end{figure}

%% file: tab/sota_occ_metrics.tex
\begin{table}[t]
\begin{center}
\setlength{\tabcolsep}{0.5mm}{
      \linespread{2}
      \begin{tabular}
      {p{0.1\textwidth}<{\centering}p{0.13\textwidth}p{0.05\textwidth}<{\centering}p{0.053\textwidth}<{\centering}p{0.053\textwidth}<{\centering}p{0.055\textwidth}<{\centering}}
         \toprule
         Type & Methods & AP & AP$_{so}$ &AP$_{mo}$ &AP$_{ho}$  \\
         \Xhline{0.6pt}
         \multirow{2}{*}{Per-frame}
         & IDOL  {\small \cite{IDOL}}              &  24.3 & 34.8 & 29.0 & 8.1 \\
         & MinVIS {\small \cite{huang2022minvis}}  &  26.3 & 45.8 & 31.6 & 9.9 \\
         \hline
         \multirow{3}{*}{Per-clip}
         & SeqFormer {\small \cite{seqformer}}  &  15.1 & 31.8 & 17.3 & 3.2 \\   
         & VITA  {\small \cite{heo2022vita}}    &  19.6 & 32.3 & 20.4 & 8.2\\
         & MDQE (ours)                          &  \textbf{29.2} & \textbf{46.6} & \textbf{33.9} & \textbf{13.1} \\
        \bottomrule
      \end{tabular}
}
\end{center}
\vspace{-5.5mm}
\caption{ Occlusion metrics of SOTA methods on OVIS valid set.}\label{tab:sota_occ_matric}
\vspace{-2.5mm}
\end{table}

%% file: tab/sota_swinl.tex
\begin{table}[t]
\begin{center}
\setlength{\tabcolsep}{0.5mm}{
      \linespread{2}
      \begin{tabular}
      {p{0.06\textwidth}p{0.13\textwidth}
      p{0.045\textwidth}<{\centering}p{0.05\textwidth}<{\centering}p{0.05\textwidth}<{\centering}
      p{0.05\textwidth}<{\centering}p{0.05\textwidth}<{\centering}}
        \toprule
         Data\ & Methods &\ AP\  &\ AP$_{50}$\ &\ AP$_{75}$\ &\ AR$_{1}$\ &\ AR$_{10}$\ \\
        \midrule
         \multirow{5}{*}{\shortstack{YT21}} 
         & IDOL {\small \cite{IDOL}}                    & 56.1 & 80.8 & 63.5 & 45.0  &60.1 \\
         & MinVIS {\small \cite{huang2022minvis}}       & 55.3 & 76.6 & 62.0 & 45.9 & 60.8 \\
         & SeqFromer{\small \cite{seqformer}}           & 51.8 & 74.6 & 58.2 & 42.8 & 58.1 \\
         & VITA {\small \cite{heo2022vita}}             & 57.5 & 80.6 & 61.0 & 47.7 & 62.6 \\
         & MDQE (ours)                                   & 56.2 & 80.0 & 61.1 & 44.9 & 59.1 \\
        \midrule
         \multirow{5}{*}{\shortstack{OVIS}} 
         & MinVIS {\small \cite{huang2022minvis}}    & 41.6 & 65.4 & 43.4 & 18.6 & 44.9\\
         & VITA {\small \cite{heo2022vita}}          & 27.7 & 51.9 & 24.9 & 14.9 & 33.0\\
         & MDQE (ours)                                & 41.0 & 67.9 & 42.7 & 18.3 & 45.2\\
         \cline{2-7}
         & IDOL $\dag${\small \cite{IDOL}}           & 42.6 & 65.7 & 45.2 & 17.9 & 49.6 \\
         & MDQE $\dag$ (ours)                         & 42.6 & 67.8 & 44.3 & 18.3 & 46.5 \\ 
         \bottomrule
      \end{tabular}
}
\end{center}
\vspace{-5.5mm}
\caption{ Quantitative performance comparison of VIS methods with Swin Large (SwinL) backbone \cite{vaswani2017attention} on benchmark VIS datasets. Symbol `$^\dag$' means that the input video is of 720p.
}\label{tab:sota_swinl}
\vspace{-2.5mm}
\end{table}

%% file: sec/5_conclusion.tex
\section{Conclusions}
\vspace{-1mm}

We proposed to mine discriminative query embeddings to segment occluded instances on challenging videos. We first initialized the positional and content embeddings of frame-level queries by considering instance spatio-temporal features. We then performed contrastive learning on instance embeddings by proposing a new inter-instance mask repulsion loss.
The proposed per-clip VIS method, termed as MDQE, was validated on OVIS and YouTube-VIS datasets. The experimental results showed that the mined discriminative embeddings of instance queries can teach the query-based segmenter to better distinguish occluded instances in crowded scenes, improving significantly performance on the challenging OVIS dataset. 


%% file: fig/query_init_visual.tex
\begin{figure*}[h]
     \centering
     \includegraphics[width=0.99\textwidth]{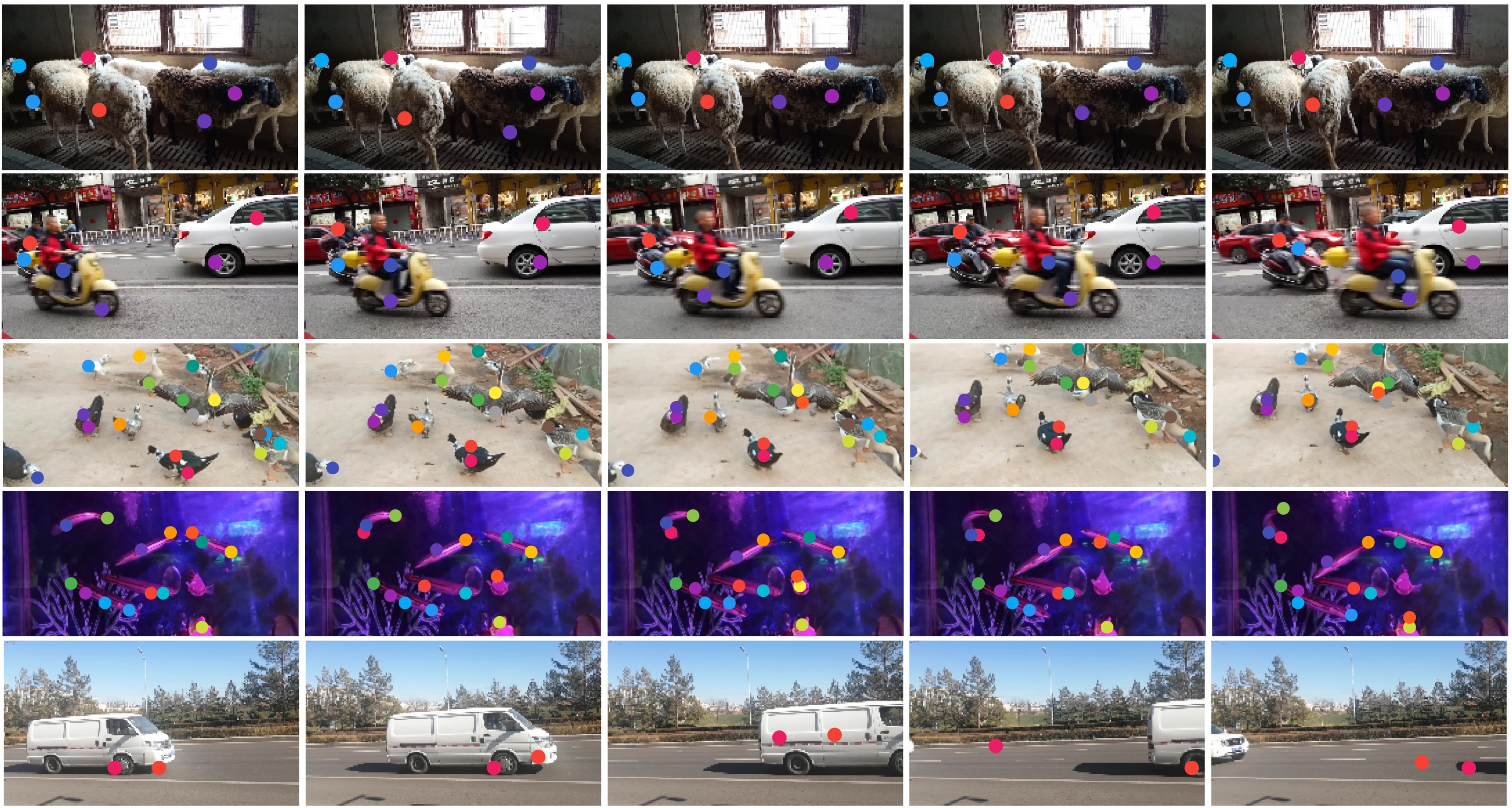}
\caption{ {\small Visualization of object query locations by our proposed grid-guided query selection and inter-frame query association method, where the highlight dots with the same color represent the initialized locations of object queries on the same object.}}\label{fig:query_init_visual}
\end{figure*}

%% file: tab/sota_yt19.tex
\begin{table*}[t]
\begin{center}
\renewcommand\arraystretch{1.1}
\setlength{\tabcolsep}{0.5mm}{
      \linespread{2}
      \begin{tabular}
      {p{0.12\textwidth}<{\centering}p{0.17\textwidth}|
      p{0.055\textwidth}<{\centering}p{0.055\textwidth}<{\centering}p{0.055\textwidth}<{\centering}p{0.055\textwidth}<{\centering}p{0.055\textwidth}<{\centering}|p{0.06\textwidth}<{\centering}p{0.055\textwidth}<{\centering}p{0.055\textwidth}<{\centering}p{0.055\textwidth}<{\centering}p{0.06\textwidth}<{\centering}}
         \Xhline{1pt}
         
         \multirow{2}{*}{Type} & \multirow{2}{*}{Method} & \multicolumn{5}{c|}{ResNet 50} & \multicolumn{5}{c}{ResNet 101}  \\
          & &AP &AP$_{50}$ &AP$_{75}$ &AR$_{1}$ &AR$_{10}$ &AP&AP$_{50}$&AP$_{75}$&AR$_{1}$&AR$_{10}$  \\
         \Xhline{1pt}
         \multirow{7}{*}{\shortstack{Per-frame}}
         &MaskTrack \cite{yang2019video}\             & 30.3 & 51.1 & 32.6  &31.0 &35.5 & 31.8 & 53.0  & 33.6 & 33.2 & 37.6  \\
         &SipMask \cite{cao2020sipmask}              & 32.5 & 53.0 & 33.3  &33.5 &38.9 & 35.0 &  56.1 & 35.2  & 36.0 &41.2  \\
         &STMask\cite{Li_2021_CVPR}                  & 33.5 & 52.1 & 36.9  &31.1 &39.2 & 36.8 &  56.8 & 38.0 & 34.8 & 41.8  \\
         &SG-Net\cite{liu2021sg}                     & 34.8 & 56.1 & 36.8 & 35.8 & 40.8 & 36.3 & 57.1 & 39.6 & 35.9 & 43.0 \\
         &CrossVIS \cite{yang2021crossover}          & 34.8 & 54.6 & 37.9 & 34.0 &39.0 & 36.6 & 57.3 & 39.7 & 36.0 & 42.0  \\ 
         & IDOL \cite{IDOL}                          & 49.5 & 74.0 & 52.9 & 47.7 &  58.7  & 50.1 & 73.1 & 56.1 & 47.0 & 57.9  \\
         & MinVIS \cite{huang2022minvis}             & 47.4 & 69.0 & 52.1 & 45.7   & 55.7 & - & - & - & - & - \\
         
         \Xhline{0.5pt}
         \multirow{9}{*}{\shortstack{Per-clip}} 
         &STEm-Seg \cite{Athar_Mahadevan20stemseg}  & 30.6 & 50.7 & 33.5 & 31.6 & 37.1 & 34.6 & 55.8 & 37.9 & 34.4 & 41.6  \\
         &MaskProp \cite{bertasius2020classifying}  & 40.0 &  -   & 42.9  &-    &- & 42.5 &- & 45.6 & - & -\\
         &Propose-Reduce \cite{lin2021video}        & 40.4 & 63.0 & 43.8 & 41.1 & 49.7 & 43.8 & 65.5 & 47.4 & 43.0 & 53.2 \\ 
         &VisTR \cite{wang2020vistr}                & 35.6 & 56.8 & 37.0 & 35.2 & 40.2 & 40.1 & 64.0 & 45.0 & 38.3 & 44.9  \\
         &EfficentVIS \cite{wu2022trackletquery}    & 37.9 & 59.7 & 43.0 & 40.3 & 46.6 & 39.8 & 61.8 & 44.7 & 42.1 & 49.8 \\
         &IFC\cite{hwang2021video}                  & 41.0 & 62.1 & 45.4 & 43.5 & 52.7 & 42.6 & 66.6 & 46.3 & 43.5 & 51.4 \\
         &SeqFromer\cite{seqformer}                 & 47.4 & 69.8 & 51.8 & 45.5 & 54.8 & 49.0 & 71.1 & 55.7 & 46.8 & 56.9\\
         &VITA \cite{heo2022vita}                   & 49.8 & 72.6 & 54.5 & 49.4 & 61.0 & 51.9 & 75.4 & 57.0 & 49.6 & 59.1 \\
         &MDQE (ours)                               & 47.3 & 66.9 & 53.1 & 42.9 & 52.9 & 47.9 & 70.3 & 53.8 & 43.2 & 53.1\\
         
         \Xhline{1pt}

      \end{tabular}
}
\end{center}
\vspace{-3mm}
\caption{ {\small {Quantitative performance comparison of methods with ResNet50 and ResNet101 on YouTube-VIS 2019 valid set. }}
}\label{tab:sota_yt19}
\vspace{-2mm}
\end{table*}

%% file: fig/yt21_visual.tex
\begin{figure*}[t]
     \centering
     \includegraphics[width=0.95\textwidth]{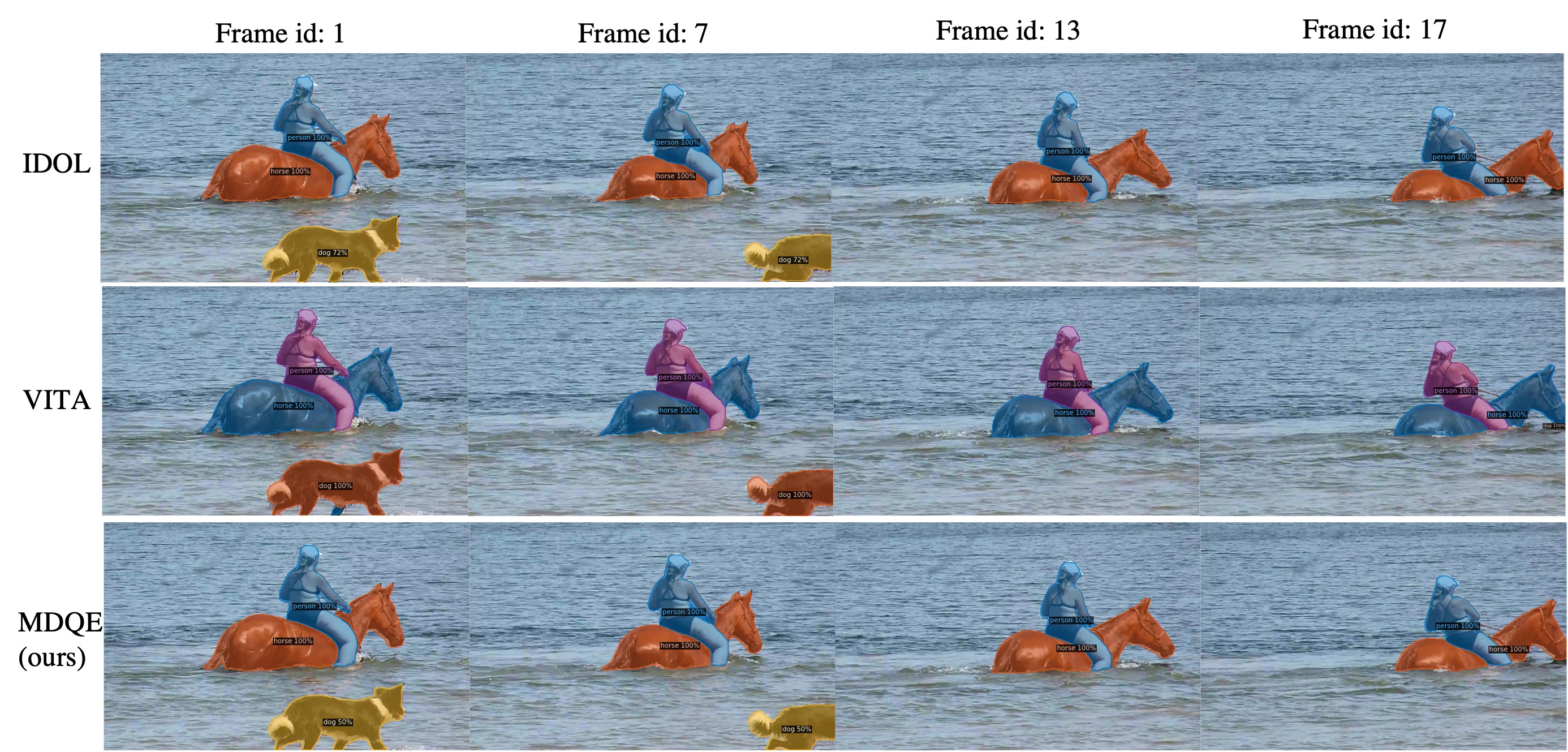}
     \includegraphics[width=0.95\textwidth]{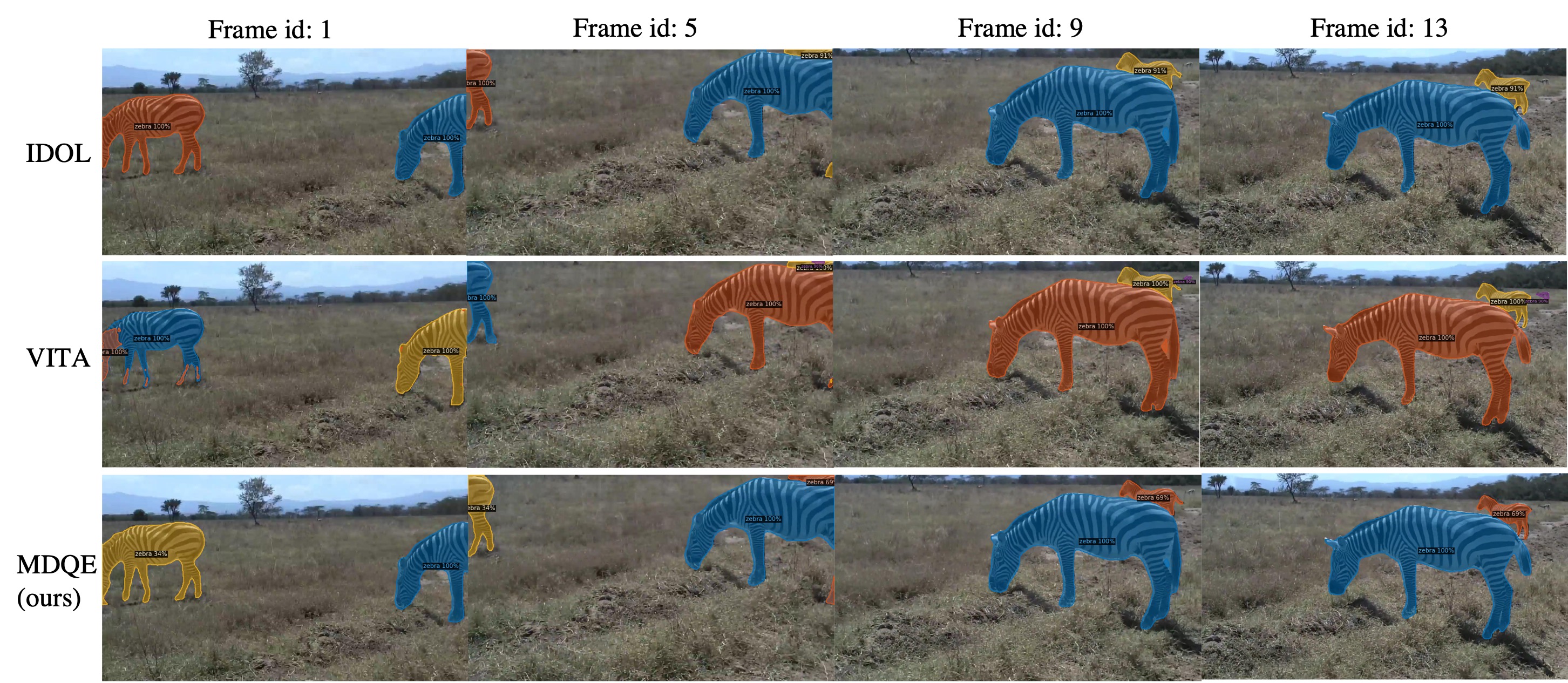}
     \includegraphics[width=0.95\textwidth]{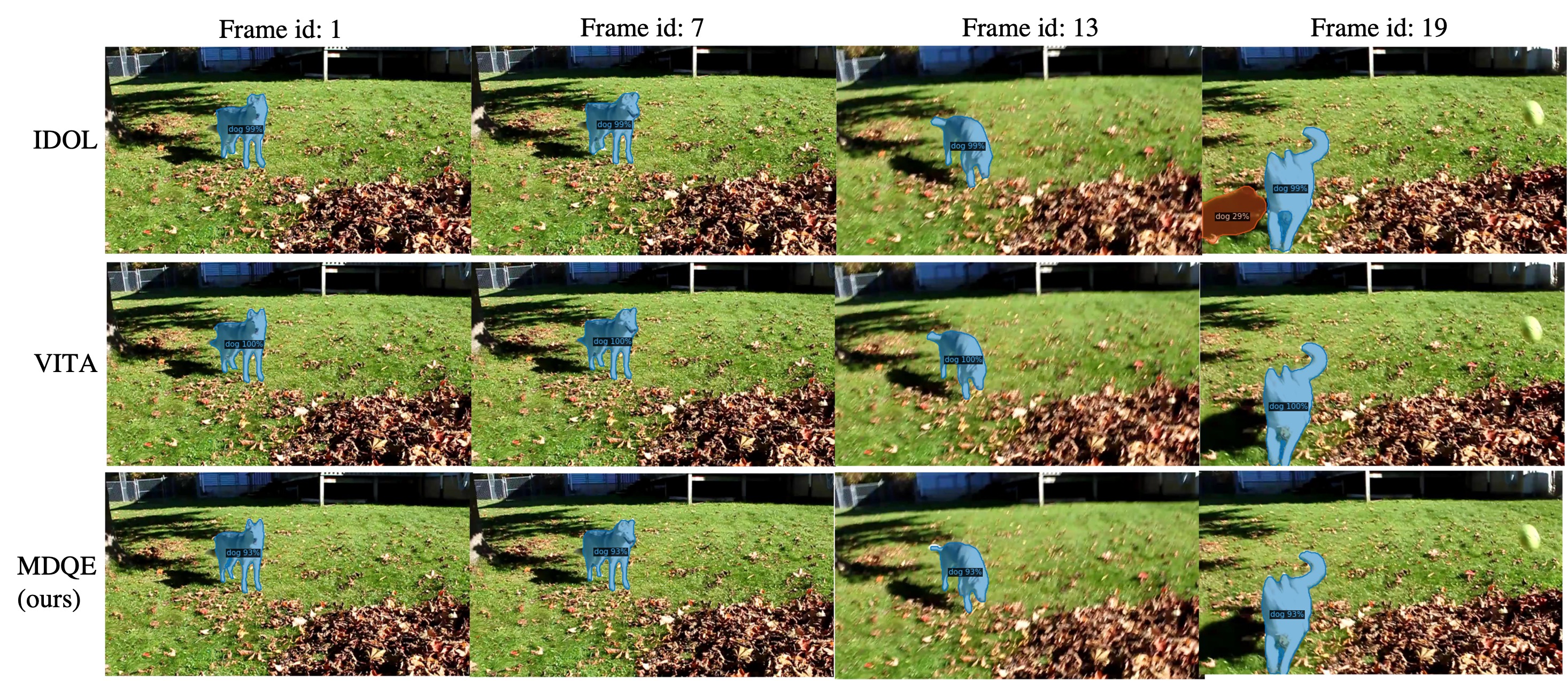}
     \vspace{-2mm}
\caption{ {\small Visualization of instance masks on videos from the YouTube-VIS 2021 valid set.}}\label{fig:yt21_visual}
\end{figure*}

%% file: fig/ovis_visual.tex
\begin{figure*}[t]
     \centering
     \includegraphics[width=0.95\textwidth]{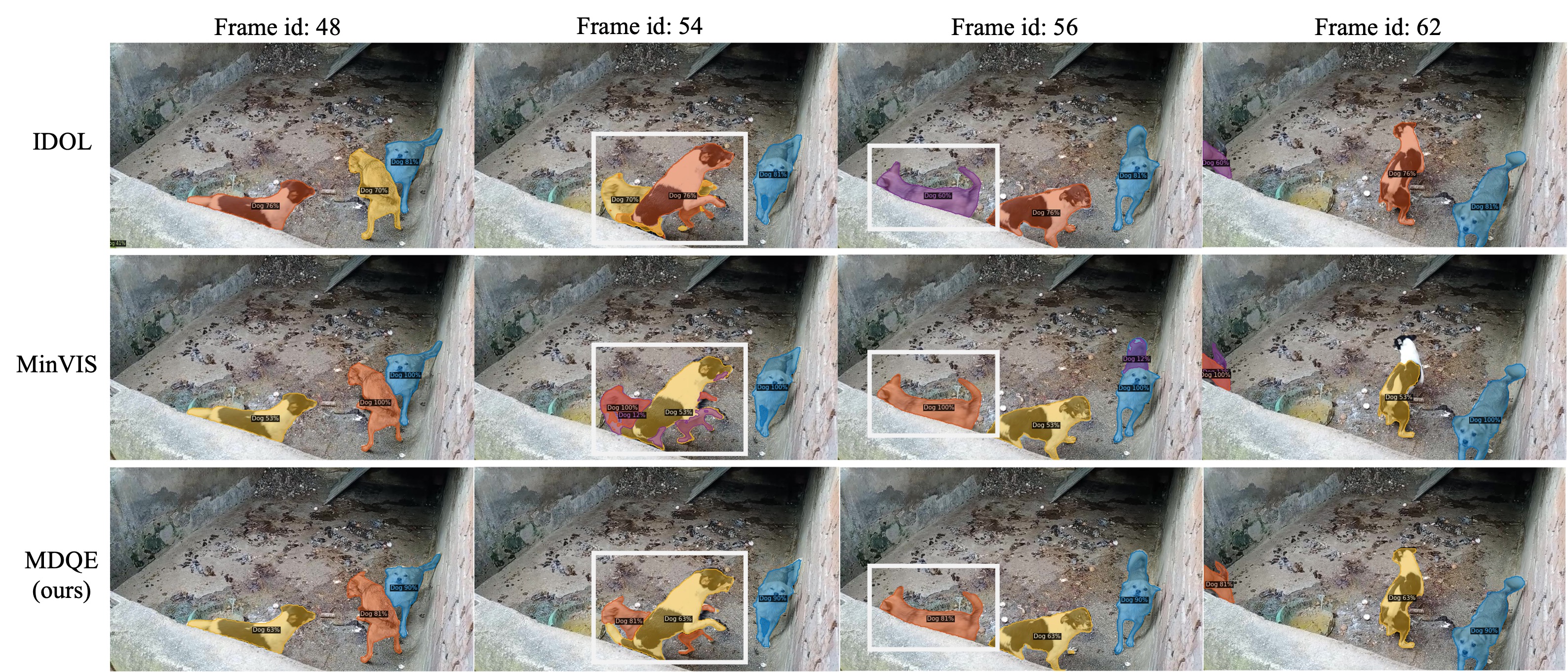}
     \includegraphics[width=0.95\textwidth]{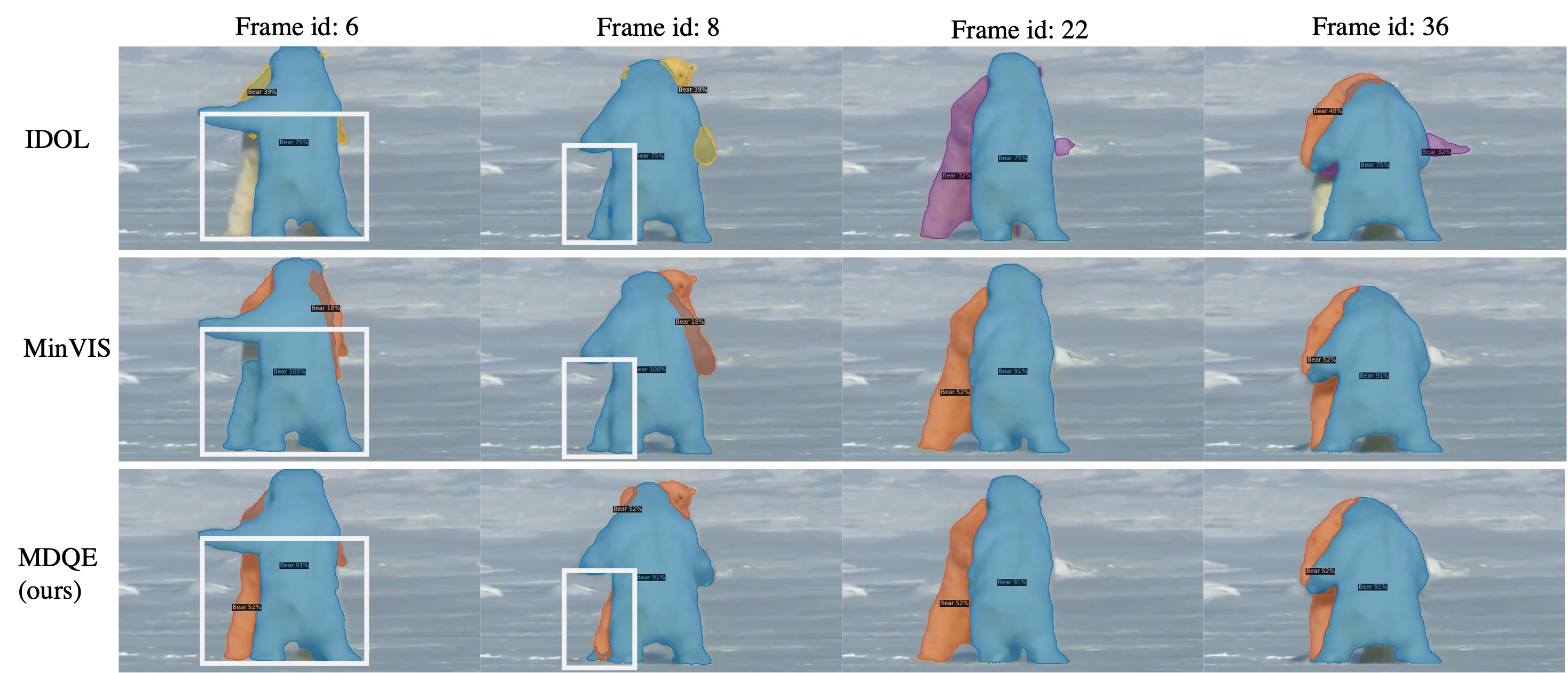}
     \includegraphics[width=0.95\textwidth]{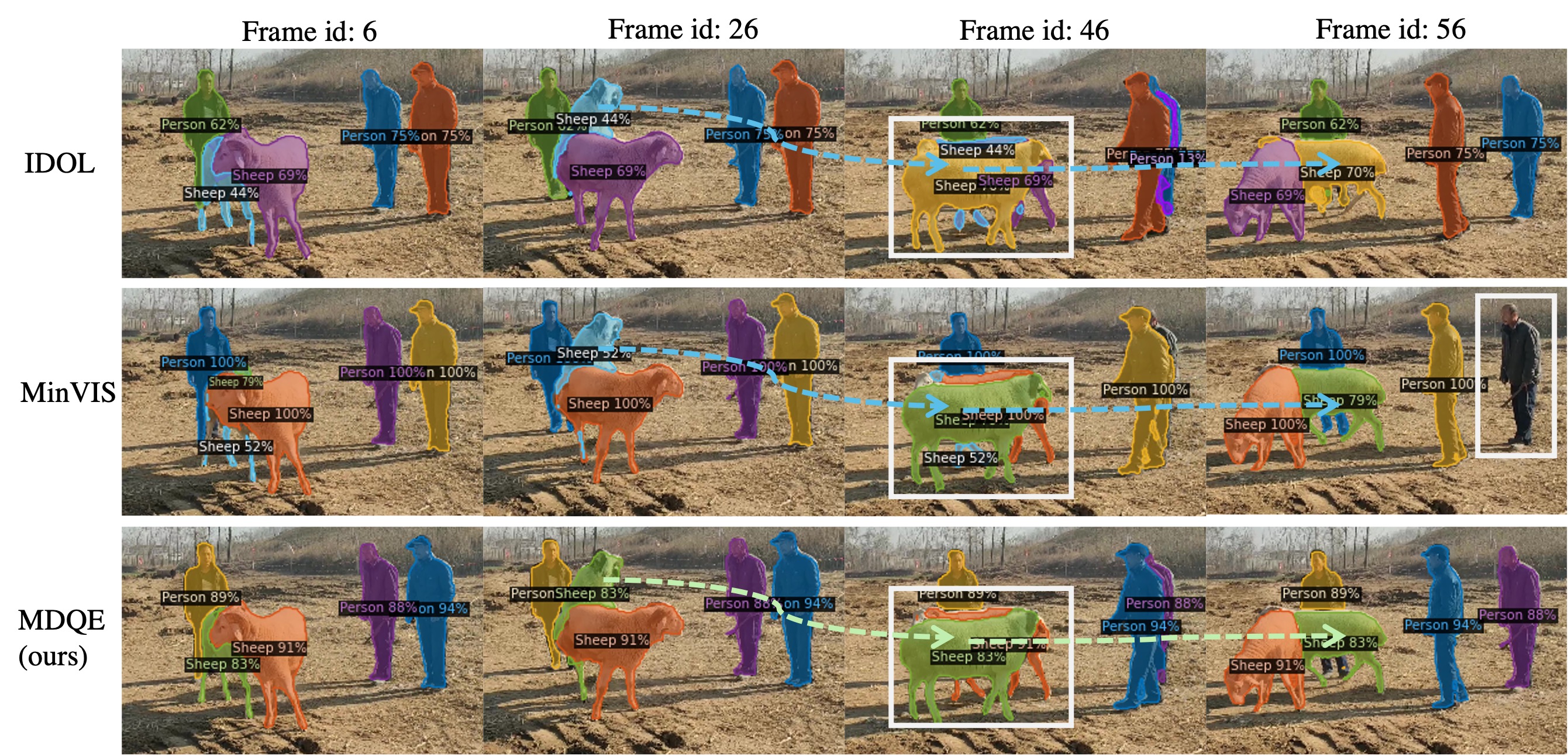}
     \vspace{-2mm}
\caption{ {\small Visualization of instance masks on the OVIS valid set.}}\label{fig:ovis_visual}
\vspace{-2mm}
\end{figure*}

%% file: fig/ovis_visual2.tex
\begin{figure*}[t]
     \centering
     \includegraphics[width=0.95\textwidth]{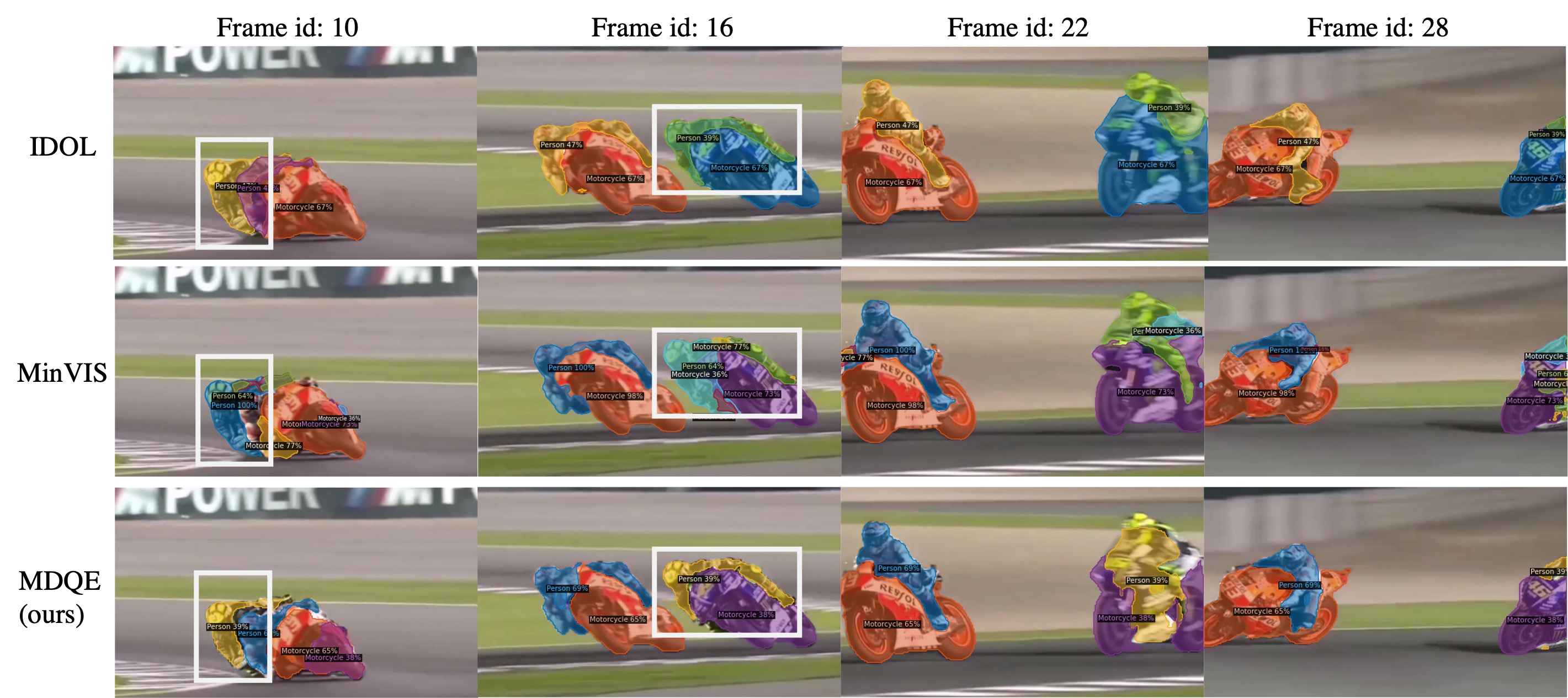}
     \includegraphics[width=0.95\textwidth]{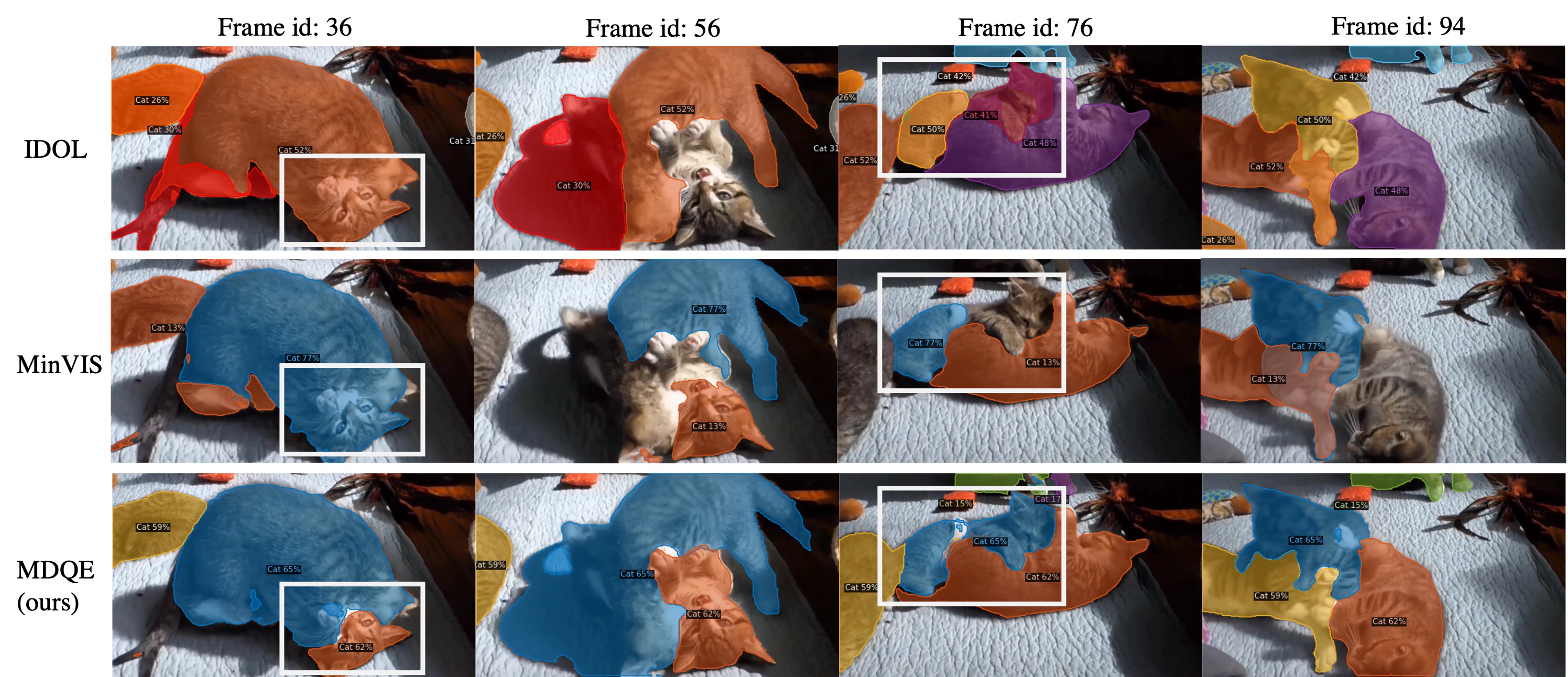}
     \includegraphics[width=0.95\textwidth]{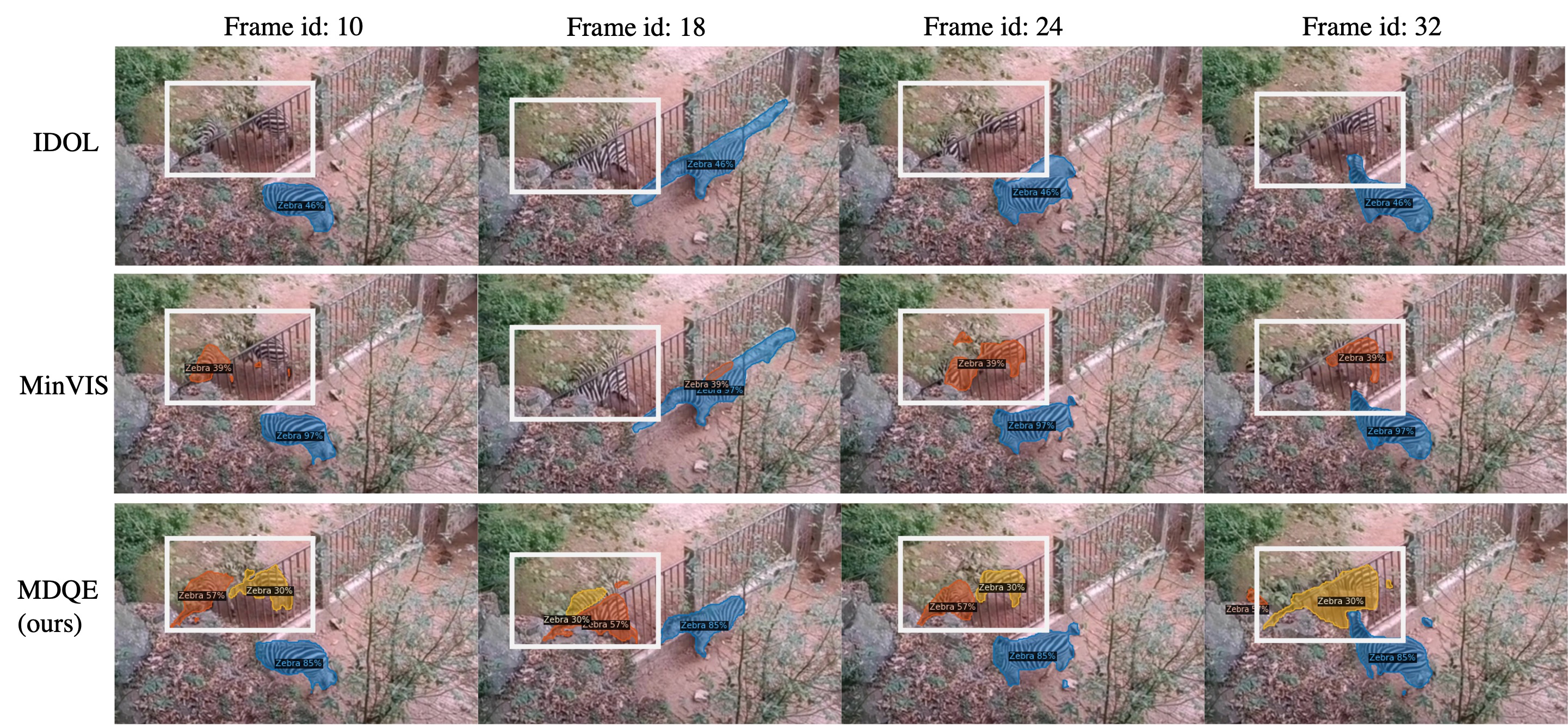}
     \vspace{-2mm}
\caption{ {\small Visualization of instance masks on the OVIS valid set.}}\label{fig:ovis_visual2}
\vspace{-2mm}
\end{figure*}